\pdfoutput=1
\PassOptionsToPackage{hyphens}{url}
\documentclass{article}
\usepackage{iclr2027_conference,times}

\usepackage{amsmath,amsfonts,bm}

\def\eqref#1{equation~\ref{#1}}

\def\1{\bm{1}}

\DeclareMathAlphabet{\mathsfit}{\encodingdefault}{\sfdefault}{m}{sl}
\SetMathAlphabet{\mathsfit}{bold}{\encodingdefault}{\sfdefault}{bx}{n}

\usepackage{hyperref}
\usepackage{url}
\usepackage{graphicx}
\usepackage{booktabs}
\usepackage{amsmath,amssymb}
\usepackage{multirow}
\usepackage{amsmath,amssymb}
\usepackage{graphicx}
\usepackage{algorithm}
\usepackage{subcaption}
\usepackage{wrapfig}
\usepackage{xcolor}
\usepackage{pifont}
\usepackage{tikz}
\usepackage{algorithm}
\usepackage{algpseudocode}
\usepackage{booktabs}
\usepackage{multirow}
\usepackage{array}
\usepackage{placeins}
\usepackage{hyperref}

\definecolor{linkblue}{RGB}{70,130,180}

\hypersetup{
    colorlinks=true,
    urlcolor=linkblue,
    citecolor=linkblue,
    linkcolor=linkblue
}
\newcommand{\tpfgain}[2]{#1~{\textcolor{red}{\scriptsize$\uparrow$\,#2}}}
\usepackage{xcolor}
\usepackage{makecell}
\title{On-Policy Residual Learning for Diffusion Language Models}

\newcommand{\orlcodeurl}{https://github.com/Introspective-Diffusion/ORL}
\newcommand{\orlwebsiteurl}{https://introspective-diffusion.github.io/orl-dlm}
\newcommand{\orlmodelsurl}{https://huggingface.co/collections/yifanyu/orl-on-policy-residual-learning-for-diffusion-lms}
\newcommand{\orlengineurl}{https://github.com/Introspective-Diffusion/DLMRT}

\definecolor{orlghblack}{RGB}{36,41,47}
\definecolor{orlwebblue}{RGB}{26,115,232}
\definecolor{orlenginegreen}{RGB}{46,164,79}
\definecolor{orllabel}{RGB}{40,40,40}
\newcommand{\orllink}[3]{\href{#1}{#2\hspace{0.32em}\textcolor{orllabel}{\textbf{#3}}}}
\newcommand{\orllinksep}{\hspace{0.9em}\textcolor{black!22}{\rule[-0.35ex]{0.5pt}{2.3ex}}\hspace{0.9em}}
\newcommand{\orlprinturl}[1]{\expandafter\url\expandafter{#1}}

\author{%
Yifan Yu\textsuperscript{1,2,*},
Yuqing Jian\textsuperscript{1,*},
Jisen Li\textsuperscript{1,*},
Yuezhou Hu\textsuperscript{3},
Tianyi Zhang\textsuperscript{1}, \\
\bfseries Xiaoxia Wu\textsuperscript{1},
Rui-Jie Zhu\textsuperscript{4},
Junxiong Wang\textsuperscript{1},
Zhongzhu Zhou\textsuperscript{1,$\dagger$}, \\
\bfseries Fan Lai\textsuperscript{2,$\dagger$,$\ddagger$},
James Zou\textsuperscript{1,5,$\dagger$},
Chenfeng Xu\textsuperscript{1,6,$\dagger$,$\ddagger$} \\
\textsuperscript{1}Together AI \\
\textsuperscript{2}University of Illinois Urbana-Champaign \\
\textsuperscript{3}University of California, Berkeley \\
\textsuperscript{4}University of California, Santa Cruz \\
\textsuperscript{5}Stanford University \\
\textsuperscript{6}The University of Texas at Austin \\[2pt]
\textsuperscript{*}Equal contribution.
\textsuperscript{$\dagger$}Equal advising.
\textsuperscript{$\ddagger$}Corresponding authors.
}

\newif\ifarxivhighlight
\arxivhighlightfalse
\definecolor{arxivred}{RGB}{226,90,90}

\providecommand{\ORLTBD}{\textnormal{TBD}}
\newcommand{\dlmrt}{\textsc{dlmrt}}

\usepackage{graphicx}
\usepackage{fontawesome5}
\usepackage{twemojis}
\usepackage[inline]{enumitem}
\newenvironment{denseitemize}{
\begin{itemize}[topsep=0pt, partopsep=0pt, leftmargin=1.5em]
  \setlength{\itemsep}{2pt}
  \setlength{\parskip}{0pt}
  \setlength{\parsep}{0pt}
}{\end{itemize}}
\iclrfinalcopy
\begin{document}

\maketitle

\begin{abstract}

Residual context \citep{team2026diffusiongemma}
is emerging as a key ingredient for unlocking the speed–quality Pareto frontier of diffusion language models (DLMs) by carrying forward predictive distributions across denoising steps. Yet turning this seemingly simple mechanism of computation recirculation into an effective and efficient learning paradigm is far from simple: there is no clear recipe for how residual context should be learned. We introduce ORL, a framework for on-policy residual learning in DLMs. Algorithmically, we identify that residual learning is inherently on-policy. ORL learns directly from these model-induced residual trajectories on the fly and introduces an entropy-contraction objective that encourages sharpening predictions as the model revisits its own intermediate predictions. System-wise, exact residual supervision requires capturing, retaining, and repeatedly processing many trajectory-dependent intermediate states. ORL co-designs the learning algorithm with asynchronous residual capture and transport, specialized block-diffusion attention, and packed compute-balanced training, making on-policy residual learning practical at scale. On a 4B DLM converted from Qwen3-4B, ORL improves tokens per forward (TPF) by up to 6.2× and, with our proposed \dlmrt{} engine, end-to-end throughput by approximately 6.9× at comparable or better quality. In single-request serving on one H100, this model achieves a stronger speed–quality Pareto frontier than autoregressive models accelerated with EAGLE3 and DFlash speculative decoding under both strict and relaxed verification, running 1.74× faster than DFlash at matched accuracy. ORL also transfers to existing DLMs: on LLaDA-2.1-mini, 36 H100-hours and 4.2M self-generated tokens, about 10× less data than prior LLaDA recipes, yield up to 1.5× higher TPF and up to 3× higher throughput at comparable quality, and on DiffusionGemma, ORL raises TPF by 1.1–1.2×. With \dlmrt{} and NVFP4 quantization, ORL models decode a single request at up to 5,046 tokens/s on one B200, 7,708 tokens/s on eight B200s, and 2,788 tokens/s on a consumer RTX 5090, with comparable quality.

\end{abstract}

\begin{center}
\orllink{\orlcodeurl}{\textcolor{orlghblack}{\faGithub}}{Code}\orllinksep
\orllink{\orlwebsiteurl}{\textcolor{orlwebblue}{\faGlobeAmericas}}{Website}\orllinksep
\orllink{\orlmodelsurl}{\raisebox{-0.17ex}{\twemoji[height=1.0em]{1f917}}}{Models}\orllinksep
\orllink{\orlengineurl}{\textcolor{orlenginegreen}{\faBolt}}{Engine}
\end{center}

\section{Introduction}
\label{sec:intro}
\suppressfloats[t]

Today’s LLMs collapse the complexity of language generation into an extremely simple next-token prediction (NTP) paradigm. This simplicity delivers remarkable intelligence, but at a fundamental systems cost: autoregressive (AR) decoding remains sequential, severely underutilizing the massive parallelism of modern hardware~\citep{nebius2025inference,frye2026speculative,robbins2026productioninference,killinger2025gpuutilization}. Unlocking parallel decoding for LLMs has emerged as one of the most timely problems in efficient language generation~\citep{almeida2026jev,liu2024deepseek,yu2026introspective,bie2026llada2}. Among the most actively explored paradigms are discrete diffusion language models (DLMs). By initializing blocks of masked tokens and progressively unmasking them, DLMs transform each forward pass from a single-token prediction into an opportunity to decode many tokens~\citep{nie2025large,arriola2025block,ye2025dream,cheng2025sdar}.

\begin{figure*}[t]
\centering
\begin{minipage}[c]{0.485\textwidth}
\centering
\includegraphics[width=\linewidth]{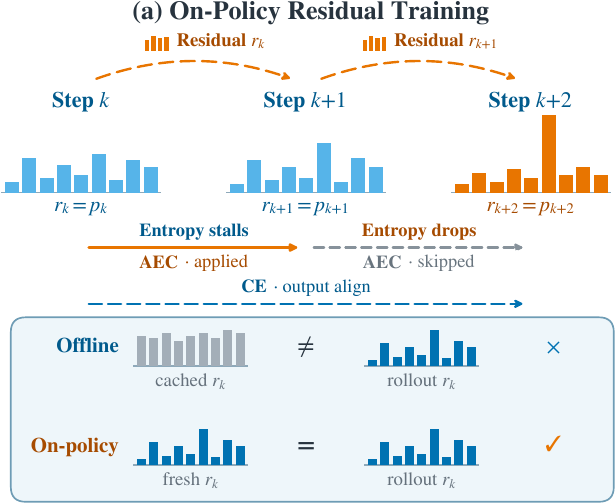}
\end{minipage}\hfill
\begin{minipage}[c]{0.4995\textwidth}
\centering
\includegraphics[width=\linewidth]{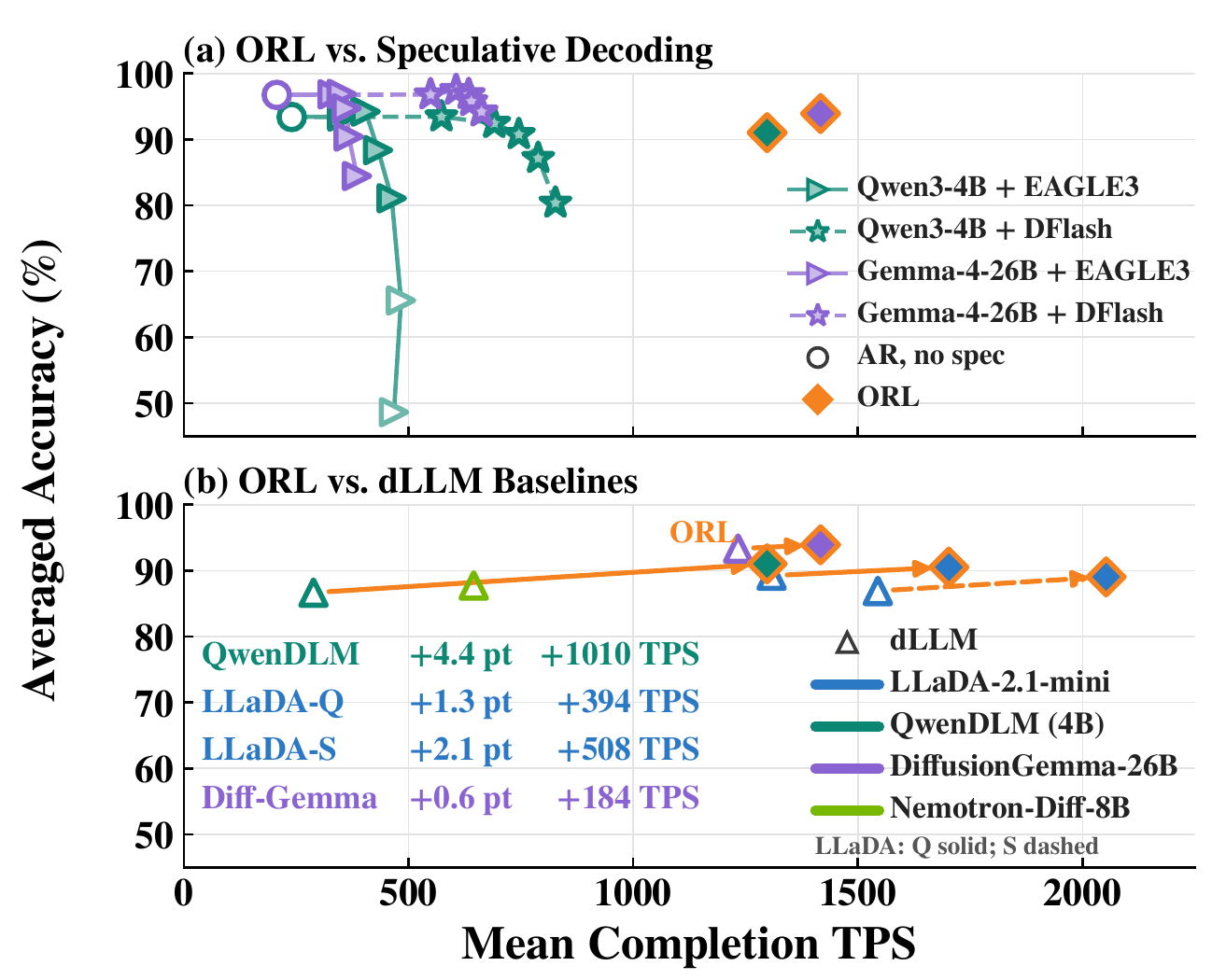}
\end{minipage}
\caption{\textbf{(a)} On-policy residual training. Residual predictions $r_k$ are carried across denoising steps. \textsc{ORL} trains on the fresh residuals produced by the current rollout (on-policy) rather than on cached residuals that no longer match the rollout (offline), and enforces entropy contraction across steps, applied only where entropy stalls, so that more tokens are committed. \textbf{(b)} A DLM target that learns to refine its residuals reaches the speculative-decoding frontier. Average GSM8K/HumanEval quality. }
\label{fig:teaser}
\end{figure*}

Recently, there has also been an interesting debate: given speculative decoding, a lossless acceleration technique, why do we still need diffusion language models? We point out that DLMs open a different optimization dimension for language generation: parallelism that the model itself learns. By contrast, existing LLMs via next-token prediction optimize next-token accuracy. Speculative decoding~\citep{chen2026dflashblockdiffusionflash,cheng2026dspark,huang2026domino} poses another dimension: it optimizes speed for a fixed target, and the attainable speedup is capped by how predictable the target already is. DLMs instead expose quality and parallelism as a joint learning problem: the model must learn not only what to predict but also how much reliable progress to make in each forward pass. Optimizing both dimensions in a single model is what makes a better speed–quality tradeoff possible than acceleration around a fixed target. It is one unique advantage of a DLM target (Figure~\ref{fig:teaser}b).

Yet optimizing the speed–quality tradeoff is difficult: beyond deciding which token should fill each position, the model must also determine how aggressively it should fill the block. Existing work tackles this challenge through improved diffusion objectives, decoding strategies, reinforcement learning, and iterative self-refinement~\citep{hu2026residual,team2026diffusiongemma,lee2026looped,wang2026revolutionizing,yu2026introspective,hu2026lightningrl,ren2026cforce}.
Among these directions, one idea is emerging as particularly natural and effective: recirculating residual predictions across decoding steps. In particular, each forward pass already spends substantial computation constructing a predictive distribution over the unresolved block. Rather than discarding this distribution and recomputing from scratch at the next step, residual decoding feeds it back as context, allowing subsequent steps to build upon the model’s previous predictions~\citep{hu2026residual,jo2026loopholing,lee2026looped}. Appendix~\ref{app:residual_learning_diagnostics} shows that these intermediate predictions often anticipate the final tokens before they are committed.
RCD \citep{hu2026residual} and Looped Diffusion \citep{lee2026looped} show that this residual information can significantly improve the speed--quality Pareto frontier. DiffusionGemma~\citep{team2026diffusiongemma} pushes this trend even further, achieving over 20 tokens per forward (TPF; about 1,230 tokens/s on one H100, as shown in Figure~\ref{fig:teaser}b), significantly faster than existing speculative decoding.

Existing methods, however, largely treat residual information as an architectural or decoding primitive, leaving no clear recipe for learning it effectively and efficiently. Algorithmically, residual learning is hard for two reasons. First, the learning loop is self-referential: Offline training cannot faithfully reproduce the states seen at inference time, namely partially denoised blocks together with the residuals they receive. Even for on-policy training, under a shared optimizer, the residual pathway moves much faster than the backbone, and naïve updates destabilize training. Second, supervising intermediate states toward the final output indicates what each state should predict, but not how much progress each step should make. In practice, we find that entropy stays nearly flat across denoising steps, and TPF barely improves (Figure~\ref{fig:entropy_refinement}). On the systems side, exact residual supervision requires generating intermediate predictions in flight and repeatedly feeding these states back to training. This trajectory-dependent workflow fits poorly into existing RL pipelines (\S\ref{sec:motivation}).

We introduce \textsc{ORL}, an on-policy residual learning framework for DLMs that addresses each of these challenges. \textbf{(1)} Residual learning is inherently on-policy, and the loop it closes must be kept stable. \textsc{ORL} learns from residual trajectories generated online by the current model, using the exact residual inputs encountered during rollout (Figure~\ref{fig:teaser}a, bottom). It rescales the residual pathway so that it updates at the backbone's pace and gates supervision by rollout entropy, ensuring that already-confident predictions are not sharpened further. \textbf{(2)} Refinement speed needs its own objective. \textsc{ORL} introduces an adjacent entropy-contraction objective that explicitly requires each step to reduce the uncertainty left by the previous one, sharpening predictions along the residual trajectory (Figure~\ref{fig:teaser}a, top). On the systems side, \textbf{(3)} \textsc{ORL} makes exact on-policy residual learning practical at scale. It asynchronously captures residual states off the decoding critical path, uses memory-adaptive transport to keep or offload them depending on placement, and packs trajectory-dependent states into dense training batches. A compute-aware scheduler balances the resulting irregular workloads across GPUs, while a specialized staircase attention kernel executes packed block-diffusion states without materializing dense masks. Together, these designs make residual training efficient, while our proposed single-request engine, \dlmrt{}, further converts fewer forward passes into higher inference throughput (Section~\ref{sec:method_inference}).

\noindent\textbf{Contributions.}
To the best of our knowledge, \textsc{ORL} represents the first open, end-to-end workflow for exact on-policy residual optimization in DLMs, and we release code, models, training data, the \dlmrt{} inference engine, and a project page to facilitate reproduction and follow-up work. Our contributions are:

\begin{denseitemize}
\item \textbf{We identify residual refinement as an inherently on-policy learning problem.}
Existing residual training suffers from two mismatches: offline residuals can induce trajectories different from those encountered at inference, while pre-generated trajectories become stale as the model evolves. This reveals that residuals must be optimized on the evolving state distribution they induce, with the residual pathway updated slowly enough that the loop stays stable.

\item \textbf{We introduce a general recipe for on-policy residual optimization.}
\textsc{ORL} trains intermediate states from fresh rollouts using their exact residual conditioning, distilling them toward verified final outputs, and explicitly improves step-wise refinement through adjacent entropy contraction with stabilized residual updates. The same recipe supports direct post-training of existing DLMs, without requiring a model-specific training pipeline.

\item \textbf{We make exact trajectory-level residual optimization practical at scale.}
We co-design the rollout and training system, including asynchronous in-engine trajectory capture, memory-adaptive residual transport, specialized block-diffusion attention, packed intermediate-state training, and compute-aware GPU balancing. These techniques speed up training by up to $5.2\times$. 

\item \textbf{We establish a stronger speed--quality frontier with little post-training.}
With 60 H100-hours of post-training on QwenDLM, a 4B DLM converted from Qwen3-4B, \textsc{ORL} improves TPF by $2.5$--$6.2\times$ and throughput by approximately $5\times$ at comparable quality (approximately $6.9\times$ together with \dlmrt{}); in single-request serving on one H100, the regime speculative decoding targets, the resulting model is faster than every EAGLE3~\citep{li2025eagle} and DFlash setting we measure: $1.74\times$ faster than DFlash at matched accuracy with relaxed verification, and $2.26\times$ faster than DFlash with strict verification, which keeps the AR's accuracy, $2.4$ points above ours (Figure~\ref{fig:teaser}b).
The same recipe improves LLaDA-2.1-mini  by up to $1.5\times$ higher TPF and up to $3\times$ higher throughput, and DiffusionGemma by $1.1$--$1.2\times$ higher TPF. \textsc{ORL} achieves these benefits with a few million self-generated tokens: on LLaDA-2.1-mini it uses 4.2M tokens and 36 H100-hours, about $10\times$ less data than the roughly 40M tokens of d3LLM-LLaDA and dParallel (Figure~\ref{fig:data_perf}; Appendix~\ref{app:training-compute}).

\item \textbf{We design \dlmrt{}, a low-latency single-request engine for block-diffusion DLMs.}
Many practical setups, such as edge AI, serve one request at a time, where engines built for batching leave the GPU underused.
Running each block's denoising loop on the GPU as one CUDA graph, \dlmrt{} raises throughput by $1.2$--$1.6\times$ over SGLang and vLLM without changing the decoding rule. With NVFP4 quantization, LLaDA-2.1-mini + \textsc{ORL} averages $5{,}046$ tokens/s per single request across HumanEval on one B200, and DiffusionGemma + \textsc{ORL} averages $4{,}214$ tokens/s across GSM8K on one B200 and $7{,}708$ tokens/s with tensor parallelism across eight B200s. On a consumer GeForce RTX 5090, the two models reach $2{,}788$ and $1{,}540$ tokens/s on these benchmarks, respectively, matching or exceeding BF16 on an H100 (Section~\ref{sec:deployment}).

\end{denseitemize}

\section{Related Work and Motivation}
\label{sec:motivation}
\label{sec:residual_decoding}

\textbf{Residual diffusion decoding.}
A block DLM generates a response block by block: each block of $B$ tokens is denoised over several steps, conditioned on the prompt and the previously committed blocks, which we call its \emph{clean context}. At step $k$, the backbone $f_\theta$ observes a discrete canvas
$s^{(k)}=(x^{(k)},m^{(k)})$, where $x^{(k)}\in\mathcal V^B$ holds the current tokens and $m^{(k)}\in\{0,1\}^B$ marks the unresolved positions, which hold noise tokens (a mask token or random vocabulary tokens), and predicts a distribution $p_i^{(k)}$ over the vocabulary $\mathcal V$ for every position. A decoding rule then accepts the unresolved positions whose predictions are confident enough, for example under a probability threshold or an entropy budget, may edit accepted ones, and re-noises the rest; the block is committed once no position remains unresolved or a step cap is reached. The number of steps $K$ of a block therefore depends on the model and the decoding rule, and TPF counts completion tokens per denoising forward. Residual decoding additionally carries the previous step's predictions forward
as continuous context~\citep{hu2026residual,jo2026loopholing,team2026diffusiongemma}. Let $r_i^{(k)}$ denote the residual at position $i$ and $E_v$ the input embedding of token $v$.
We inject its expected token embedding through a lightweight residual adapter $g_\phi$ with injection strength $\alpha$:
\begin{equation}
\widetilde e_i^{(k)} = E_{x_i^{(k)}} + \alpha\, g_\phi\big(z_i^{(k)}\big),
\qquad z_i^{(k)}=\textstyle\sum_{v\in\mathcal V} r_{i,v}^{(k)}E_v,
\qquad k>1.
\label{eq:residual_input}
\end{equation}
Here $r^{(1)}=0$, and each position receives the distribution it produced at the previous step. For backbones with next-token (shifted) alignment, such as QwenDLM, the output at position $i$ predicts position $i{+}1$, so $r_i^{(k+1)}=p_{i+1}^{(k)}$; same-position backbones, such as LLaDA, use $r_i^{(k+1)}=p_i^{(k)}$. The adapter is model-specific: QwenDLM uses a gated MLP warmed up offline, LLaDA a zero-initialized gated MLP added for \textsc{ORL}, and DiffusionGemma its native residual module (Appendix~\ref{app:optimization_stabilization}).
We call $\tau_{\theta,\phi}
=\{(s^{(k)},p^{(k)},r^{(k)})\}_{k=1}^{K}$
a \emph{residual trajectory}. We call the pair $(s^{(k)},r^{(k)})$, a canvas together with the residual it receives, the \emph{state} at step $k$, and index the states of a response by block and step, $(b,k)$. The adapter and the residual it injects form the \emph{residual pathway}.

\begin{figure*}[t]
\centering
\begin{subfigure}[t]{0.32\textwidth}
\centering
\vspace{0pt}
\includegraphics[width=\linewidth]{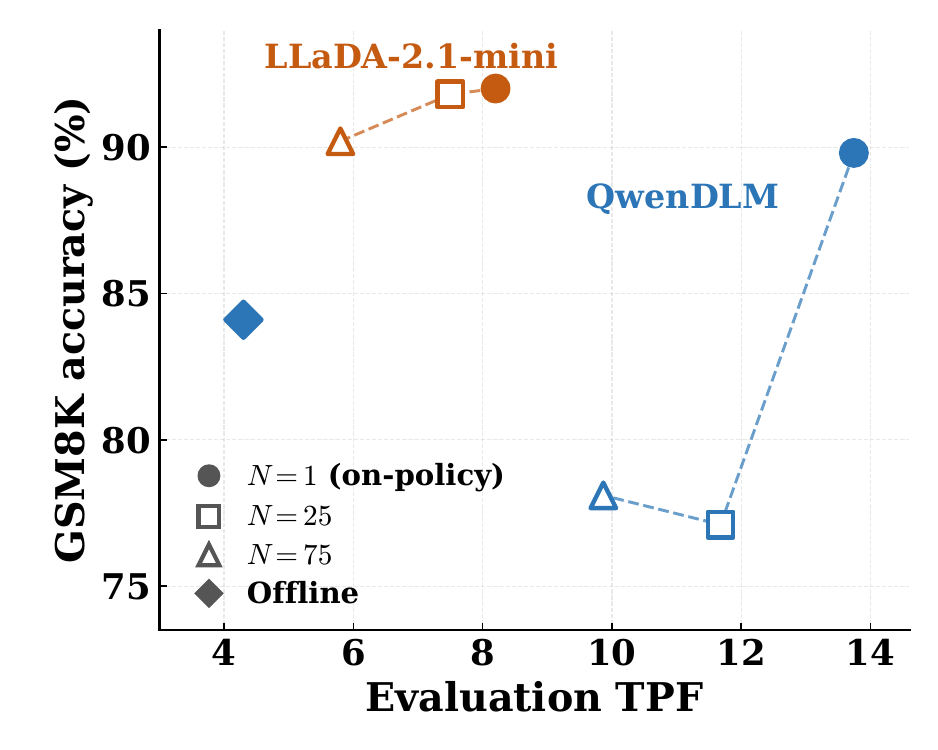}
\caption{\textbf{Residual learning is on-policy. 
}}
\label{fig:motivation_a}
\label{fig:onpolicy_ablation}
\end{subfigure}\hfill
\begin{subfigure}[t]{0.32\textwidth}
\centering
\vspace{0pt}
\includegraphics[width=\linewidth]{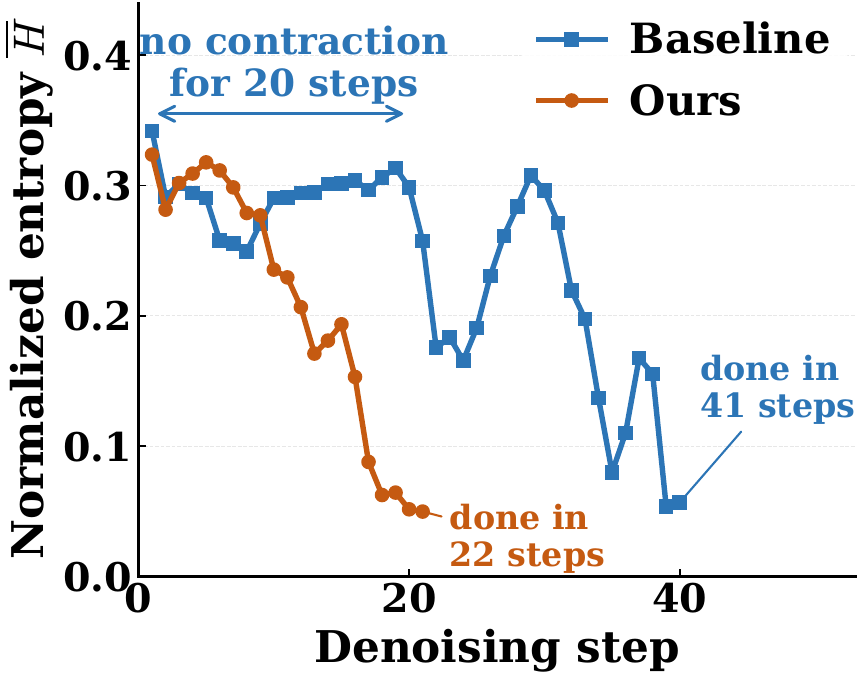}
\caption{\textbf{Accuracy alone does not accelerate refinement.}}
\label{fig:entropy_refinement}
\end{subfigure}\hfill
\begin{subfigure}[t]{0.32\textwidth}
\centering
\vspace{0pt}
\includegraphics[width=\linewidth]{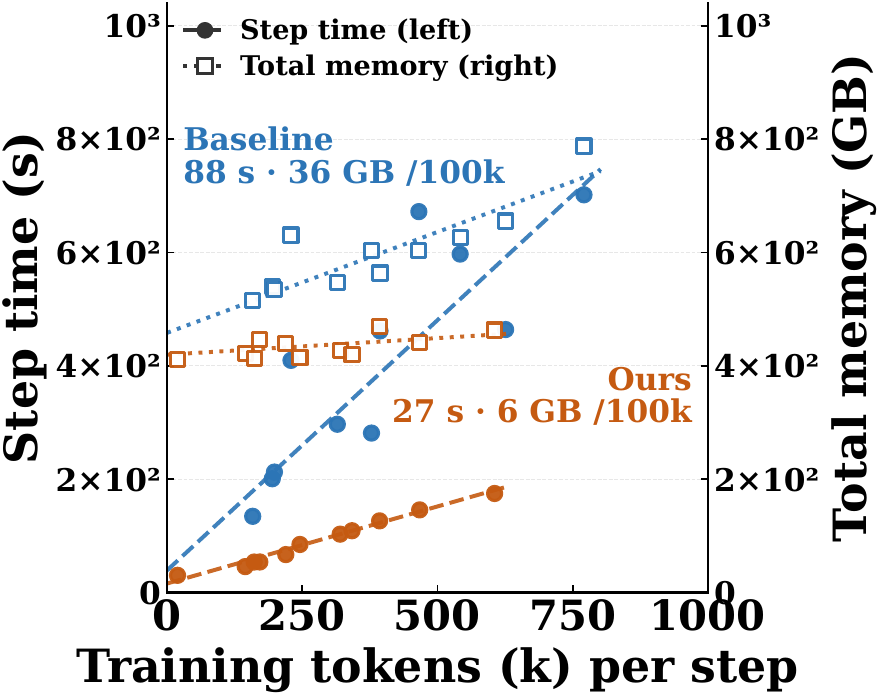}
\caption{\textbf{Exact on-policy training is systems-heavy.}}
\label{fig:motivation_c}
\end{subfigure}
\vspace{-.2cm}
\caption{
\textbf{Three observations behind \textsc{ORL}.}
\textbf{(a)} 
Only fresh trajectories jointly improve accuracy and TPF.
\textbf{(b)} Standard residual supervision leaves entropy nearly flat across
denoising steps.
\textbf{(c)} Exact on-policy training incurs substantial time and memory overhead.
}
\label{fig:motivation}
\end{figure*}

\textbf{Residual learning is inherently on-policy.}
Residual predictions directly affect which tokens are committed, which positions remain unresolved, and therefore which states the model encounters next. As the model changes, the residuals it produces and the trajectories induced by those residuals change together. This makes offline and stale residual supervision fundamentally mismatched. Offline residuals~\citep{jo2026loopholing,hu2026residual}, which are not taken from the current model's own previous denoising step, may induce states different from those encountered by the model at inference time, while pre-generated trajectories~\citep{ren2026cforce} become stale as optimization
updates the model. 
We define staleness $N$ as optimizer steps between rollout-policy
syncs ($N{=}1$ is on-policy; offline uses fixed cached trajectories).
Existing on-policy DLM methods instead optimize token-space trajectories or final
outcomes without directly learning the residual pathway
~\citep{wang2026revolutionizing,hu2026lightningrl}.
Figure~\ref{fig:motivation}(a) shows that offline training barely improves TPF,
while increasing staleness sacrifices accuracy; only fresh trajectories improve both. Section~\ref{sec:entropy_dynamics} shows that training keeps changing the model's behavior, with rollout entropy decreasing steadily without collapse, so trajectories from earlier checkpoints become stale.

\textbf{Accuracy supervision does not imply faster refinement.}
On-policy trajectories determine \emph{where} residual learning should happen, but not \emph{what behavior} it should encourage.
Standard supervision trains each state toward the correct output, yet provides no pressure for one denoising step to make more progress than the previous one. In practice, we observe that the average normalized entropy of the unresolved positions (Section~\ref{sec:onpolicy_residual}) remains nearly flat across many
steps, even when the final answer is correct (Figure~\ref{fig:motivation}b). The model learns to predict accurately, but not to become progressively more decisive.
For parallel decoding, this distinction is critical: improving final accuracy does not necessarily increase the number of tokens that can be committed per forward pass. Residual learning, therefore, needs an explicit objective for \emph{refinement
speed}, not accuracy alone.

\textbf{Exact residual learning creates a new systems workload.}
Correct on-policy supervision requires training on the exact states seen during rollouts. These states are generated dynamically at every denoising step, must cross the
rollout–training boundary without perturbing generation, and produce examples with highly irregular trajectory lengths and compute costs. This workload fits poorly into conventional RL training pipelines~\citep{sheng2024hybridflow,fu2025areal}.
A naïve implementation repeatedly materializes these states, wastes memory on sparse trajectory structures, and suffers severe load imbalance. As Figure~\ref{fig:motivation}(c) and Appendix~\ref{app:motivation_system_scaling} show, its time and memory costs grow rapidly. Thus, \textbf{on-policy residual learning requires a system around the states of residual
trajectories.}

Appendix~\ref{app:related_work} discusses additional related work.

\section{Method}
\label{sec:method}

\begin{figure}[t]
    \centering
    \includegraphics[width=\linewidth]{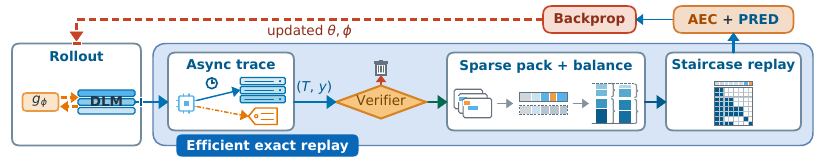}
    \caption{
        \textbf{Overview of \textsc{ORL}.} The current model generates residual-conditioned trajectories with exact state capture; verified states are packed and load-balanced across GPUs, replayed with staircase attention, and optimized using CE and AEC.
    }
    \label{fig:system_overview}
\end{figure}

Previous research discussed in Section~\ref{sec:motivation} drew three main conclusions: residuals must be learned on the trajectories the current model induces (Figure~\ref{fig:motivation}a), refinement speed needs its own objective (Figure~\ref{fig:motivation}b), and exact on-policy supervision is a systems workload in its own right (Figure~\ref{fig:motivation}c). \textsc{ORL} answers all three within a loop (Figure~\ref{fig:system_overview}). In each iteration, the model generates fresh residual-conditioned trajectories, a verifier keeps the successful rollouts, and \textsc{ORL} replays their states, re-evaluating each with the exact residual input seen during generation. Two objectives are applied: verified-output supervision teaches \emph{what} each state should predict, and adjacent entropy contraction teaches \emph{how quickly} predictions should sharpen across denoising steps. As every update changes the conditioning that generates the next batch, we control how fast the residual pathway moves (Section~\ref{sec:onpolicy_residual}). Section~\ref{sec:method_systems} describes the training system that makes exact replay practical, and Section~\ref{sec:method_inference} the inference optimizations that turn fewer forward passes into higher throughput.

\subsection{On-Policy Residual Learning}
\label{sec:onpolicy_residual}
\label{sec:refinement_objective}

Because residuals and the states they induce change together with the model (Figure~\ref{fig:motivation}a), \textsc{ORL} regenerates its training data at every optimization step. At optimization step $u$, the model $(\theta_u,\phi_u)$ generates $G$ rollouts for each prompt $c$ in minibatch $\mathcal B$. Each rollout $\mathcal R$ yields a final response $y^{\mathcal R}$ and its residual-conditioned trajectory. A verifier $V(c,y)\in\{0,1\}$ keeps the successful rollouts $\mathcal R^{+}_{\mathcal B}$; rejected rollouts contribute no gradient. Each nonempty verified batch yields one optimizer step; its trajectories are then discarded, and the updated model generates the next batch. Within each kept rollout we train a subset $\mathcal S(\mathcal R)$ of its states; for QwenDLM, for example, these are the 25\% of states in each block that accept the most new tokens, plus the block's final state (Appendix~\ref{app:training_details}).

Equation~\ref{eq:residual_input} indicates that every step is conditioned on the previous step's prediction, so the canvas alone does not identify a state, and recovering the residual actually consumed would mean re-executing the earlier steps. \textsc{ORL} therefore preserves the \emph{exact states} that the current model visited. For each selected state $(b,k)$, we capture its canvas and the pre-adapter residual $z_i^{(b,k)}$ it consumed, computed with the rollout embedding table $E^{(u)}$. Replay evaluates Equation~\ref{eq:residual_input} on the captured canvas with $z_i^{(b,k)}$ held fixed by a stop-gradient, $\operatorname{sg}[z_i^{(b,k)}]$, so gradients reach the backbone and the adapter but not the embeddings through $z$. Replay places the captured canvas after its clean context and runs one forward pass under block-diffusion attention (Section~\ref{sec:method_systems}), producing the replayed prediction $\widehat p_i^{(b,k)}$; model-specific normalization of the injected term, identical in rollout and replay, is omitted from Equation~\ref{eq:residual_input} (Appendix~\ref{app:optimization_stabilization}). We store $z$ rather than a vocabulary-sized distribution, and gradients update the backbone and the adapter at the captured state without backpropagating through earlier steps. Exactness here means preserving the captured input, not bitwise equality between rollout and replay outputs. Unlike experience replay in off-policy RL, each captured state is replayed once, by the parameters that generated it, and then discarded. Capturing these inputs without slowing generation is the first requirement of the training system in Section~\ref{sec:method_systems}.

\textbf{Learning what to predict.}
For every selected state of a verified rollout, we supervise the replayed prediction $\widehat p_i^{(b,k)}$ toward $y^{\mathcal R}_{b,i}$, the token at position $i$ of block $b$ in the verified response, with cross-entropy (CE), $\ell^{\mathrm{pred}}_{\mathcal R,b,k,i}=-\log \widehat p^{(b,k)}_{i,\,y^{\mathcal R}_{b,i}}$, at both accepted and unresolved positions alike, so that an intermediate prediction corrected later learns directly from its eventual value. Token losses are averaged over the valid positions of each verified response and then across responses, giving $\mathcal L_{\mathrm{pred}}$, so that every verified response carries equal weight (Appendix~\ref{app:optimization_stabilization}). Endpoint supervision alone, however, does not indicate how quickly a state should reach its prediction: a model can become more accurate while making little progress from one step to the next. This is the stall in Figure~\ref{fig:motivation}(b).

\textbf{Learning how fast to refine.}
\emph{Adjacent entropy contraction} (AEC) supplies the missing pressure by asking each step to reduce the uncertainty left by the previous one. Let $\mathcal U_{b,k}=\{i:m_i^{(b,k)}=1\}$ be the unresolved positions, $\overline H(p)=-\sum_v p_v\log p_v/\log|\mathcal V|$ the normalized token entropy, and $H_{b,k}=\tfrac{1}{|\mathcal U_{b,k}|}\sum_{i\in\mathcal U_{b,k}}\overline H(p_i^{(b,k)})$ the average unresolved entropy of a state, computed from the detached rollout distribution for the source state ($H^{\mathrm{roll}}_{b,k}$, captured during generation) and from the replayed prediction $\widehat p$ for the next state ($H^{\theta,\phi}_{b,k+1}$). For a target contraction rate $\rho\in[0,1]$,
\begin{equation}
    \ell^{\mathrm{AEC}}_{\mathcal R,b,k}
    =
    \Big[
        \operatorname{ReLU}\!\Big(
            H^{\theta,\phi}_{b,k+1}
            -
            (1-\rho)\,\operatorname{sg}\big[H^{\mathrm{roll}}_{b,k}\big]
        \Big)
    \Big]^2 .
    \label{eq:aec_pair}
\end{equation}
AEC rests on two design choices. First, it is a hinge: the loss is zero once the next state's unresolved entropy has dropped by at least the fraction $\rho$, and it penalizes only steps that fail to contract. A uniform entropy penalty would instead keep sharpening every prediction, including those that already make progress, at the risk of trading accuracy for confidence; the hinge acts only where refinement stalls (Figure~\ref{fig:teaser}a). Second, it targets the positions that remain unresolved after acceptance, since each entropy is taken over its own step's unresolved set. The only information the next step receives from the previous one is the tokens just accepted or edited and the residual carrying the previous step's predictive distribution. Thus, AEC rewards the model for turning that residual into confidence on the positions still to be decoded. Pair losses over $\mathcal P(\mathcal R)$, the within-block adjacent pairs $(b,k)$ whose next state $(b,k{+}1)$ is selected and whose two unresolved sets are nonempty, are averaged with equal weight per response into $\mathcal L_{\mathrm{AEC}}$ (Appendix~\ref{app:optimization_stabilization}), and the full objective is $\mathcal L_{\mathrm{ORL}}=\mathcal L_{\mathrm{pred}}+\lambda_{\mathrm{AEC}}\mathcal L_{\mathrm{AEC}}$ with $\lambda_{\mathrm{AEC}}\ge0$. This objective trains \emph{what} each state predicts and \emph{how efficiently} the trajectory reaches it. As shown in Figure~\ref{fig:motivation}(b), AEC cuts decoding from 41 to 22 steps with the same answer.

\textbf{Stabilizing the on-policy feedback loop.}
Because \textsc{ORL} trains on its own outputs, the loop can fail in two ways, and we guard against each. \emph{The residual pathway can outrun the backbone.} Its outputs are the inputs of the next rollout, yet under a shared optimizer its relative update is about $38\times$ the backbone's, and QwenDLM's verifier reward collapses after roughly 250 training steps (Figure~\ref{fig:alpha_schedule_keep}; Appendix~\ref{app:residual_scaling}). We therefore keep the pathway's relative update size close to the backbone's: a pathway warmed up offline is rescaled by $\gamma$ without changing its function, which under Adam shrinks its relative updates by about $\gamma$, and a pathway added from scratch starts from a zero output projection with a lower learning rate (Appendix~\ref{app:optimization_stabilization}). \emph{Supervision can reinforce what the model already believes.} Every CE target is a token the model generated itself, so at positions it already predicts confidently, CE only sharpens an existing belief, and it does so again at every iteration. We therefore apply CE only where the rollout is uncertain, with normalized entropy above a threshold $\zeta$, and keep confident positions near their rollout prediction with a capped reverse KL.

\subsection{Efficient Exact Residual Training}
\label{sec:method_systems}

This loop is simple to describe but expensive to run. Exact replay turns each response into many states that must be captured during rollout, moved across the rollout--training boundary, and re-evaluated under block-diffusion visibility with highly irregular cost (Figure~\ref{fig:motivation}(c)).
\textsc{ORL} removes these costs in the order data flows through the training loop (Figure~\ref{fig:system_overview}), with three mechanisms that change how training executes, but not which states are supervised or which objective is optimized.

\textbf{Asynchronous exact-state capture.}
The first cost is getting states out of the decode loop. Once generation advances, recovering an exact residual-conditioned input would require re-executing the preceding steps, while copying each state out synchronously would stall decoding. The engine instead snapshots the canvas, the pre-adapter residual $z$, and the rollout statistics the objectives need (the source entropies for AEC and, when the entropy gate is used, the output-head inputs from which its reference is rebuilt) at the moment they are produced, enqueues GPU-side handles, and transfers them off the decode critical path. When rollout and training are colocated, the block-local buffers holding $z$, which we call residual pages, stay GPU-resident and are handed off through CUDA-IPC rather than round-tripping through host memory (Appendix~\ref{app:async_capture}).

\textbf{Packed and cost-balanced state training.}
Captured states form the irregular workload of Section~\ref{sec:motivation}. Each verified rollout contributes a variable number of states and AEC pairs with different context lengths and residual footprints, so evaluating them one by one recomputes shared clean context and leaves GPUs idle. Following the slicing of TraceRL~\citep{wang2026revolutionizing}, a row holds a response's clean tokens once together with the states of its blocks at the same step $k$, so each state reuses the clean context before it. We pack rows from different responses as independent documents, keep residuals as sparse residual pages rather than full-sequence tensors, and assign rows to data-parallel groups in order of largest estimated cost first. Each group returns the numerators and denominators of the response-normalized losses, which are summed globally before division, so packing and redistribution leave the objective unchanged up to floating-point rounding (Appendix~\ref{app:packed_balanced}).

\textbf{Staircase attention kernel.}
Packing leaves one obstacle: each packed row mixes documents, clean context, and noisy blocks, and replay must still respect block-diffusion visibility. A noisy block attends to the prompt, the preceding clean blocks, and itself, but never to its own clean copy, which is present in the row as context for later blocks, to later blocks, or to other noisy blocks; clean queries never attend to noisy keys. A dense mask costs $O(N^2)$ storage and dispatches fully invisible tiles. Our tiled Triton kernel encodes the rule implicitly from per-document metadata, skips invisible tiles, and supports execution over packed variable-length documents with grouped-query attention (Appendix~\ref{app:staircase_kernel}). Figure~\ref{fig:system_ablation} measures the cumulative effect of the three mechanisms.

\subsection{\texorpdfstring{Efficient Residual Inference}{Efficient Residual Inference}}
\label{sec:method_inference}

Fewer forward passes do not automatically mean faster decoding. Residual conditioning adds work to every forward pass, and once a request needs fewer denoising steps, costs that do not shrink with the step count, such as prefill and per-block cache writes, take a larger share of its time (Appendix~\ref{app:inference-optimization}). We therefore optimize the decoder without changing its decoding rule. Within each model family, Base and \textsc{ORL} run on the same engine, so throughput comparisons between them reflect training rather than engine differences.

\textbf{Commit folding.}
After a block is finished, some existing decoders (e.g., the LLaDA-2.1-mini decoder) run one more forward pass on the clean block to check that no token should still be edited and to write the block's key--value cache. Following the inference engine of the LLaDA-2.1 team~\citep{bie2026llada2,antgroup2025powerup}, we fold this pass into the next block's first denoising step, so a single forward pass processes the clean previous block and the masked current block together. If the check edits a token, the current block's half is discarded and the joint pass is repeated, so every block still sees the context it would see without folding. We extend this folding to more models, including QwenDLM and the residual-conditioned \textsc{ORL} models. Saving one forward pass per block gives the largest single gain, $53\%$ higher throughput (Figure~\ref{fig:llada-inference-tps-cumulative}). CUDA graphs for prompt prefill and the residual adapter, fused acceptance kernels, metadata reuse, and gradient-free scheduling remove most of the remaining host overhead; together with commit folding, they raise the HumanEval throughput of LLaDA-2.1-mini with \textsc{ORL} by $88\%$. \dlmrt{} applies the same folding.

\textbf{A single-request engine.}
General-purpose serving engines are designed around batching many requests. However, many practical setups require efficiently serving a single request, such as edge AI on local or consumer GPUs. Single-request decoding is also where parallel decoding pays off most: at batch size~1, decoding is memory-bound and most GPU compute sits idle, which is exactly the compute a DLM spends on committing many tokens per forward; it is also the regime speculative decoding targets. Each such forward pass is small, however, so kernel launches and scheduling take a large share of its time. We therefore develop \dlmrt{}, an inference engine dedicated to single-request diffusion decoding for all three model families: it runs each block's denoising loop on the GPU as one CUDA graph and rewrites the decoding kernels around a single request (Appendix~\ref{app:dlmrt}). We release \dlmrt{} as a separate repository, and Section~\ref{sec:deployment} compares it with state-of-the-art engines.

\providecommand{\ORLNA}{\textemdash}
\providecommand{\ORLTBD}{\textnormal{TBD}}
\begin{table*}[t]
    \centering
    \caption{
    \textbf{Generation quality and parallel decoding efficiency.}
    We report benchmark scores (\%) and tokens per forward (TPF).
    Our results evaluate a single checkpoint across all benchmarks.
    Base and ORL use their selected family-specific decoders
    (Appendix~\ref{app:eval_details}), so arrows compare complete
    model--decoder systems rather than weights alone.
    Within each model-family block, the best result is \textbf{bolded} and the
    starting point for ORL is \underline{underlined}. 
    Red arrows
    indicate ORL's TPF improvement. 
    }
    \label{tab:main_quality_tpf}

    \footnotesize
    \setlength{\tabcolsep}{1pt}
    \renewcommand{\arraystretch}{1}

    \begin{tabular*}{\textwidth}{
        @{}>{\raggedright\arraybackslash}p{0.33\textwidth}
        @{\extracolsep{\fill}}*{8}{c}@{}
    }
        \toprule
        \multirow{2}{*}{\textbf{Model / training}}
        & \multicolumn{2}{c}{\textbf{GSM8K}}
        & \multicolumn{2}{c}{\textbf{MATH-500}}
        & \multicolumn{2}{c}{\textbf{HumanEval}}
        & \multicolumn{2}{c}{\textbf{MBPP}} \\
        \cmidrule(lr){2-3}
        \cmidrule(lr){4-5}
        \cmidrule(lr){6-7}
        \cmidrule(lr){8-9}
        & Acc.$\uparrow$ & TPF$\uparrow$
        & Acc.$\uparrow$ & TPF$\uparrow$
        & pass@1$\uparrow$ & TPF$\uparrow$
        & pass@1$\uparrow$ & TPF$\uparrow$ \\
        \midrule

        \multicolumn{9}{@{}l}{
            \textit{Qwen-based models and other DLM baselines}
        } \\

        d3LLM--Dream
        & 81.4 & 4.9
        & 38.2 & 3.9
        & 57.1 & 3.2
        & 55.6 & 3.0 \\

        Fast-dLLM-v2-7B
        & 77.5 & 2.2
        & 48.7 & 2.6
        & 61.7 & 2.6
        & 50.1 & 2.0 \\

        RCD--SDAR-4B-b64
        & 87.0 & \ORLNA
        & 68.4 & \ORLNA
        & \ORLNA & \ORLNA
        & \ORLNA & \ORLNA \\

        TraDo-4B (TraceRL, dynamic)
        & 90.3 & \ORLNA
        & 71.8 & \ORLNA
        & \ORLNA & \ORLNA
        & \ORLNA & \ORLNA \\

        SDAR-8B-b32
        & 88.9 & 2.9
        & 63.6 & 4.8
        & 73.5 & 2.4
        & 58.0 & 2.4 \\

        \addlinespace[3pt]

        \underline{QwenDLM (4B)}
        & 90.4 & 2.4
        & 88.0 & 2.1
        & 82.9 & 1.8
        & 79.2 & 2.1 \\

        \textbf{\quad + \textsc{ORL}}
        & \textbf{92.4} & \tpfgain{\textbf{5.9}}{+3.5}
        & \textbf{92.2} & \tpfgain{\textbf{5.8}}{+3.7}
        & \textbf{89.6} & \tpfgain{\textbf{11.1}}{+9.3}
        & \textbf{79.4} & \tpfgain{\textbf{8.0}}{+5.9} \\

        \midrule

        \multicolumn{9}{@{}l}{
            \textit{LLaDA-based models}
        } \\

        Fast-dLLM--LLaDA
        & 74.7 & 2.8
        & 30.8 & 2.0
        & 37.8 & 2.6
        & 38.6 & 2.1 \\

        dParallel--LLaDA
        & 72.6 & 5.1
        & 30.2 & 3.2
        & 39.0 & 4.9
        & 40.0 & 2.4 \\

        D2F--LLaDA
        & 73.2 & 2.9
        & 28.7 & 2.4
        & 36.6 & 2.7
        & 38.0 & 1.9 \\

        d3LLM--LLaDA
        & 73.1 & \textbf{9.1}
        & 30.4 & 5.7
        & 39.6 & 6.0
        & 40.6 & 4.2 \\

        dOPSD--LLaDA-8B-Instruct
        & 72.87 & \ORLNA
        & 36.00 & \ORLNA
        & 39.63 & \ORLNA
        & 39.54 & \ORLNA \\

        dUltra-coding
        & 81.5 & 8.4
        & 35.6 & 6.7
        & 35.9 & 7.2
        & 37.0 & 7.3 \\

        DMax-Math
        & \textbf{92.1} & 5.5
        & 75.4 & 5.9
        & \ORLNA & \ORLNA
        & \ORLNA & \ORLNA \\

        \addlinespace[4pt]

        \underline{LLaDA-2.1-mini}
        & 91.9 & 5.6
        & 84.6 & 8.2
        & 86.6 & 11.3
        & \textbf{77.5} & 7.7 \\

        \textbf{\quad + \textsc{ORL}}
        & 92.0 & \tpfgain{8.5}{+2.9}
        & \textbf{87.0} & \tpfgain{\textbf{10.2}}{+2.0}
        & \textbf{89.0} & \tpfgain{\textbf{13.1}}{+1.8}
        & 77.3 & \tpfgain{\textbf{10.3}}{+2.6} \\

        \midrule

        \multicolumn{9}{@{}l}{
            \textit{DiffusionGemma-based models}
        } \\

        \underline{DiffusionGemma}
        & 95.8 & 26.0
        & \textbf{92.3} & 23.7
        & 90.9 & 24.5
        & \textbf{87.7} & 23.1 \\

        \textbf{\quad + \textsc{ORL}}
        & \textbf{96.0} & \tpfgain{\textbf{30.0}}{+4.0}
        & \textbf{92.3} & \tpfgain{\textbf{27.4}}{+3.7}
        & \textbf{91.9} & \tpfgain{\textbf{27.9}}{+3.4}
        & 87.4 & \tpfgain{\textbf{28.7}}{+5.6} \\

        \bottomrule
    \end{tabular*}
\end{table*}

\section{Experiments}
\label{sec:experiment}
\textbf{Evaluation Methodology.}
\label{sec:experimental_setup}
We evaluate \textsc{ORL} from two popular starting points on nine benchmarks.
For \emph{AR start}, we convert Qwen3-4B-Thinking-2507~\citep{yang2025qwen3}
into a residual-conditioned DLM, termed QwenDLM, through an SDAR-style
offline training pipeline,
and continue optimization with \textsc{ORL}.
For \emph{DLM start}, we post-train LLaDA-2.1-mini and
DiffusionGemma~\citep{team2026diffusiongemma}. Full training and evaluation details are in Appendix~\ref{app:training_details}.

\subsection{Main Results}
\label{sec:main_results}

\textbf{\textsc{ORL} substantially improves both quality and parallelism.}
Table~\ref{tab:main_quality_tpf} compares \textsc{ORL} with prior DLM baselines,
including TraceRL's TraDo-4B~\citep{wang2026revolutionizing} and the
on-policy self-distillation method dOPSD~\citep{dat2026dopsd}.\footnote{Baseline results are quoted from published papers; the MATH-500 entries of rows from the d3LLM evaluation are MATH (4-shot) on the full test set. Appendix~\ref{app:baselines} lists the source of each row.}
From QwenDLM, it improves TPF by $2.5$--$6.2\times$ while preserving or improving quality.
On HumanEval, for example, \textsc{ORL} reaches $89.6$ pass@1 at $11.1$ TPF, compared with $73.5/2.4$ for SDAR-8B.
The same advantage transfers to existing DLMs: on both LLaDA-2.1-mini and DiffusionGemma, \textsc{ORL} raises TPF on all four benchmarks while maintaining or improving quality.
For example, on HumanEval, \textsc{ORL} improves LLaDA-2.1-mini from $86.6/11.3$ to $89.0/13.1$ in pass@1/TPF.
On GSM8K and MATH-500, it reaches $92.0/8.5$ and $87.0/10.2$ in accuracy/TPF, compared with $92.1/5.5$ and $75.4/5.9$ for DMax-Math~\citep{chen2026dmax}.
Appendix~\ref{app:additional_compairson} compares additional methods that are not included in Table~\ref{tab:main_quality_tpf} because they lack public checkpoints or cannot be evaluated under aligned settings.

\begin{wraptable}{R}{0.45\textwidth}
\vspace{-.4cm}
    \centering
    \scriptsize
    \setlength{\tabcolsep}{2.2pt}
    \renewcommand{\arraystretch}{1.10}
    \caption{
        \textbf{Generalization across domains.}
        Post-training of LLaDA-2.1-mini and DiffusionGemma (DG)
        across additional benchmarks.
        Each entry reports \emph{quality / TPF}.
        }
\label{tab:cross_domain_generalization}
\vspace{-.3cm}
  \begin{tabular}{lcccc}
    \toprule
    \textbf{Dataset}
    & \shortstack{\textbf{LLaDA}\\\textbf{Base}}
    & \shortstack{\textbf{LLaDA}\\\textbf{+\textsc{ORL}}}
    & \shortstack{\textbf{DG}\\\textbf{Base}}
    & \shortstack{\textbf{DG}\\\textbf{+\textsc{ORL}}} \\
    \midrule
    AIME-24
    & \textbf{50.0}/5.31
    & 48.3/\textbf{7.77}
    & 55.8/18.15
    & \textbf{56.3}/\textbf{22.02} \\
    MMLU-Pro
    & 57.7/3.60
    & \textbf{58.2}/\textbf{4.88}
    & \textbf{78.6}/18.57
    & 78.1/\textbf{22.60} \\
    GPQA-D
    & \textbf{49.5}/3.05
    & 46.1/\textbf{4.21}
    & \textbf{65.7}/17.50
    & 65.3/\textbf{21.27} \\
    LCB-v6
    & 28.0/5.90
    & \textbf{30.3}/\textbf{7.43}
    & \textbf{53.5}/19.96
    & 51.4/\textbf{24.09} \\
    IFEval
    & \textbf{84.1}/1.53
    & 82.3/\textbf{1.74}
    & \textbf{66.9}/9.90
    & \textbf{66.9}/\textbf{11.68} \\
    \bottomrule
  \end{tabular}
  \vspace{-.3cm}
\end{wraptable}
\textbf{\textsc{ORL} generalizes across models and domains.}
Benchmarks in Table~\ref{tab:main_quality_tpf} cover standard math
and code evaluations but do not establish whether the same gains
transfer broadly.
Table~\ref{tab:cross_domain_generalization} evaluates direct
post-training of LLaDA-2.1-mini and DiffusionGemma on five additional benchmarks
spanning mathematics, knowledge-intensive reasoning, code
generation, and instruction following.
Across two independently developed DLMs, \textsc{ORL} raises TPF on all
ten model--benchmark pairs ($1.26\times$ on average, up to
$1.46\times$), with little quality change.

\begin{figure}[t]
\centering
\includegraphics[width=0.9\linewidth, trim=0 188.5pt 0 0, clip]{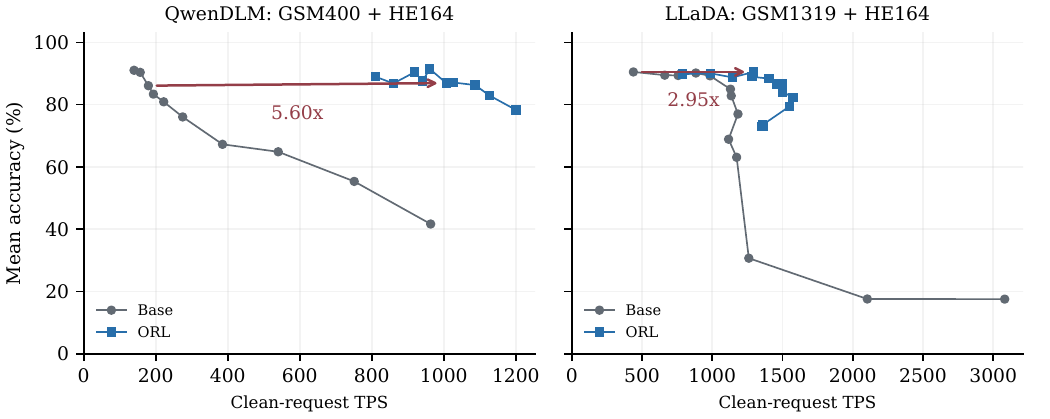}
\caption{Common-task quality--throughput sweeps: mean GSM8K/HumanEval
accuracy versus clean-request TPS for every Base and ORL setting. Arrows mark
the highlighted comparable-accuracy pairs, with $5.60\times$ (QwenDLM) and
$2.95\times$ (LLaDA-2.1-mini) clean-TPS gains.
Appendix~\ref{app:quality_throughput_frontiers} lists the settings and reports
completed-output TPS. All settings are measured with SGLang.}
\label{fig:quality_tpf_frontiers_main}
\end{figure}

\begin{figure*}[t]
    \centering
    \begin{minipage}[t]{0.68\textwidth}
        \centering
         \includegraphics[width=1\textwidth]{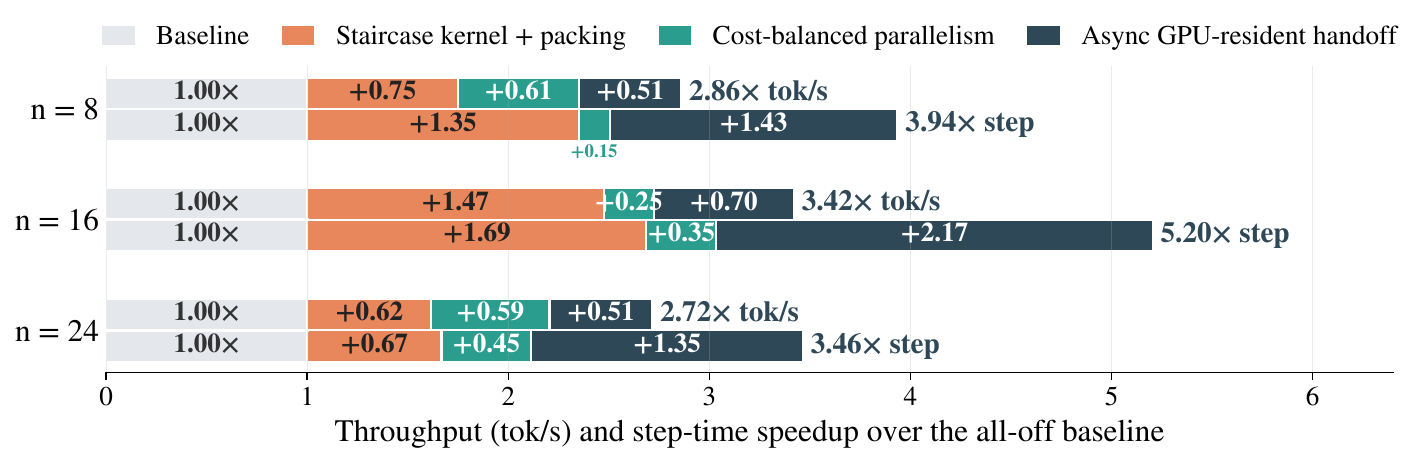}
         \vspace{-.6cm}
    \caption{
        \textbf{Cumulative system optimizations.}
        End-of-bar values report system
        speedups in training-step time and token throughput. 
    }
    \label{fig:system_ablation}
    \end{minipage}
    \hfill
    \begin{minipage}[t]{0.3\textwidth}
        \centering
        \includegraphics[width=\linewidth]{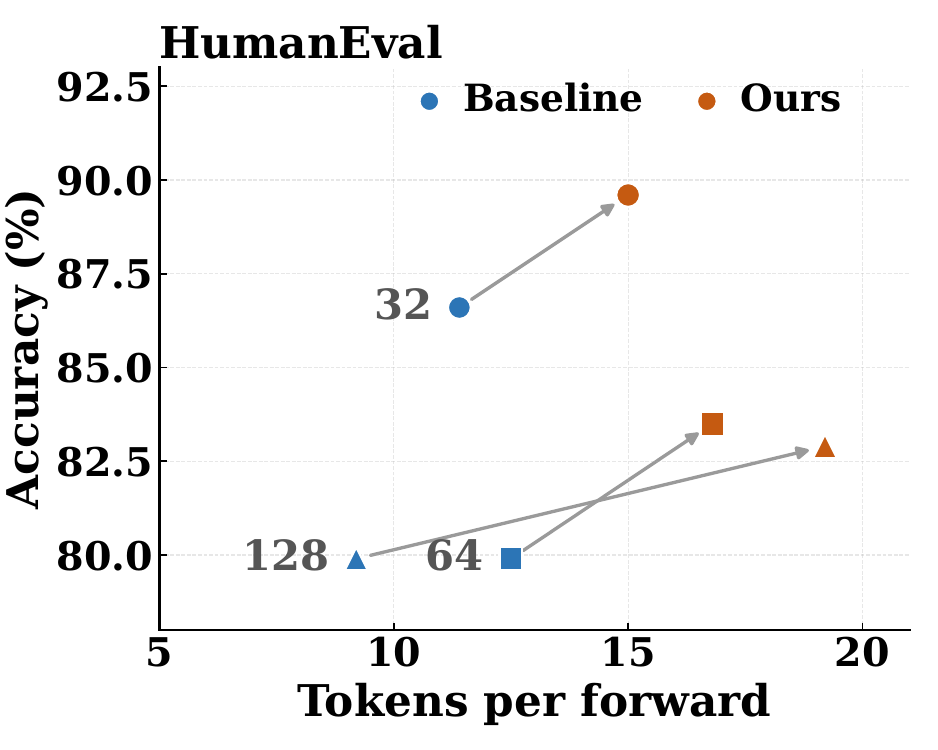}
\vspace{-.6cm}
        \captionof{figure}{
            \textbf{Scaling with training block size.}
        }
        \label{fig:block_scaling}
    \end{minipage}
\end{figure*}

\textbf{\textsc{ORL} improves the quality--throughput frontier.}
Because TPF ignores the per-forward cost of residual conditioning and
block-diffusion execution, we also measure output throughput for a single request on one
H100, running the DLMs on \dlmrt{} (Figure~\ref{fig:teaser}(b)).
QwenDLM trained with \textsc{ORL} reaches $1{,}298$ tok/s at $91.0\%$
average GSM8K/HumanEval accuracy, about $6.9\times$ faster than its starting DLM running on SGLang ($187$ tok/s), and about $5\times$ when \textsc{ORL} also runs on SGLang (Appendix~\ref{app:throughput_accounting}).
AR decoding reaches $240$ tok/s. Among measured speculative settings with
mean accuracy at least \textsc{ORL}'s minus $0.5$ percentage points, the
fastest EAGLE3~\citep{li2025eagle} and DFlash settings reach $405$ and
$746$ tok/s using relaxed verifier thresholds ($\tau{=}0.1$ and $0.03$);
with strict verification ($\tau{=}1$), they reach $353$ and $573$ tok/s.
These are system-level comparisons using each method's native decoder and
clean-request throughput, with AR and speculative decoding running on vLLM. Appendix~\ref{app:additional_compairson} reports the
full accuracy, failure-count, and verification-threshold context.
Sweeping the decoding threshold further shows that \textsc{ORL} shifts the whole
quality--throughput frontier outward rather than selecting a more aggressive
threshold, improving clean throughput by $5.60\times$ on QwenDLM
and $2.95\times$ on LLaDA-2.1-mini at the highlighted comparable-accuracy pairs
(Figure~\ref{fig:quality_tpf_frontiers_main}; Appendix~\ref{app:quality_throughput_frontiers}).
Appendix~\ref{app:tpf_curve_analysis} characterizes these accuracy--TPF curves.
Unlike speculative decoding, whose speedup is capped by how predictable a
frozen target already is, \textsc{ORL} optimizes both speed and quality, with advantage widening with block size (Figure~\ref{fig:block_scaling}).

\textbf{Rollout freshness changes learning dynamics.}
Our multi-seed study compares every-step rollout refresh ($N=1$) with
infrequent refresh ($N=25,75$), using the same initialization, verifier,
and optimization objective
(Figure~\ref{fig:onpolicy_ablation}). All settings generate new trajectories;
$N$ controls rollout-weight refresh.
After convergence, the fresh-rollout setting
reaches 89.8\% mean GSM8K accuracy and 13.7 TPF, versus 77.1\%/11.7 and
78.1\%/9.9 for $N=25$ and $N=75$, respectively.
Fresh rollouts start lower but overtake the infrequent-refresh variants.
These results motivate frequent refresh
for the evaluated recipe.

\subsection{Ablations and Analysis}
\label{sec:analysis_ablations}

\textbf{System co-design makes exact on-policy training practical.}
Figure~\ref{fig:system_ablation} cumulatively enables staircase attention
and packed state training, compute-aware balancing, and asynchronous
GPU-resident handoff.
Across $n=8/16/24$ rollouts, the complete system reduces optimizer-step time by
$3.9/5.2/3.5\times$ and raises token throughput by $2.9/3.4/2.7\times$ over the
all-off implementation.
Detailed phase breakdowns
are provided in
Appendix~\ref{app:system_analysis}.

\begin{figure*}[t]
\vspace{-.2cm}
    \centering

    \begin{minipage}[t]{0.66\textwidth}
        \centering

        \begin{minipage}[t]{0.49\linewidth}
            \centering
            \includegraphics[width=\linewidth]{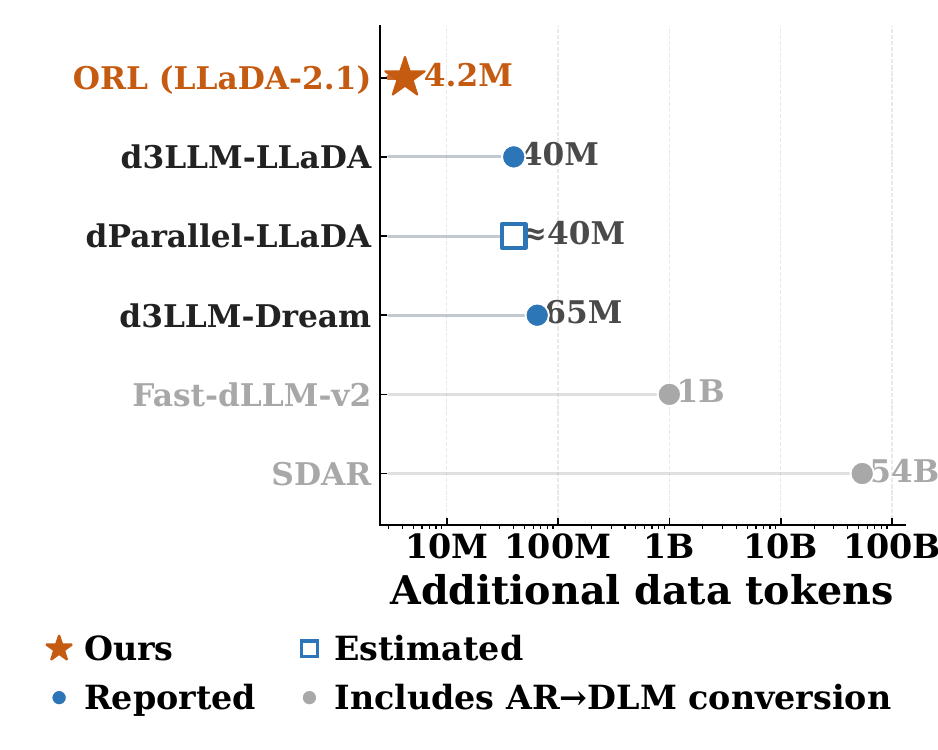}\\[2pt]
            {\small (a) Data efficiency}
        \end{minipage}
        \hfill
        \begin{minipage}[t]{0.49\linewidth}
            \centering
            \includegraphics[width=\linewidth]{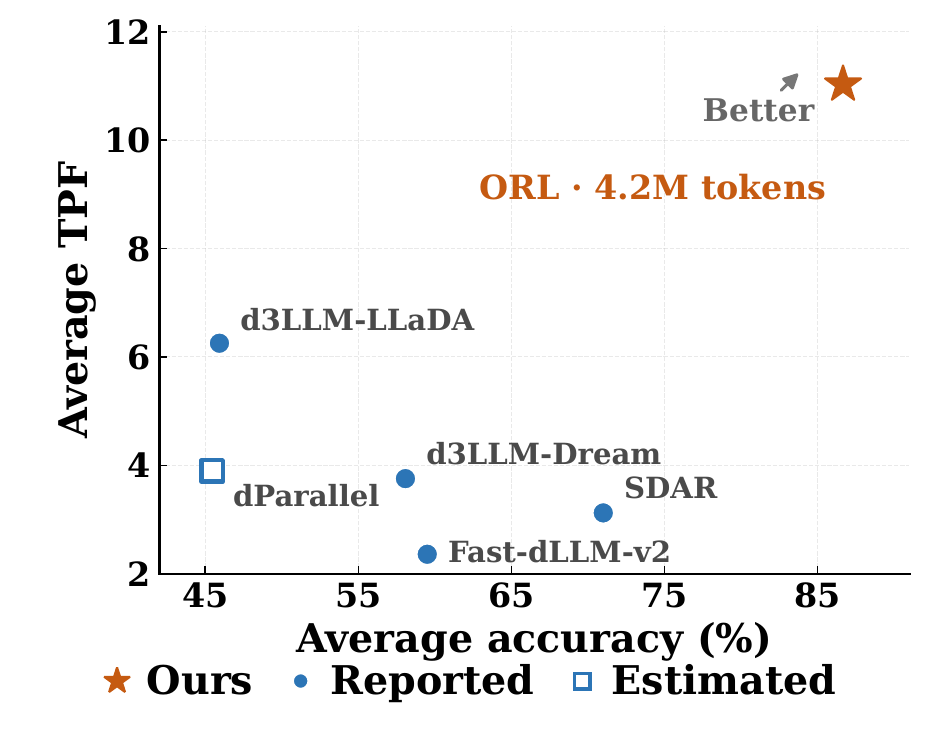}\\[2pt]
            {\small (b) Performance comparison}
        \end{minipage}
\vspace{-.2cm}
        \captionof{figure}{ \textbf{(a--b)} With only a small amount of self-generated on-policy training data, \textsc{ORL} rapidly improves refinement efficiency and achieves a stronger quality--TPF tradeoff.
        }
        \label{fig:data_perf}
    \end{minipage}
    \hfill
    \begin{minipage}[t]{0.32\textwidth}
        \centering
        \includegraphics[width=\linewidth]{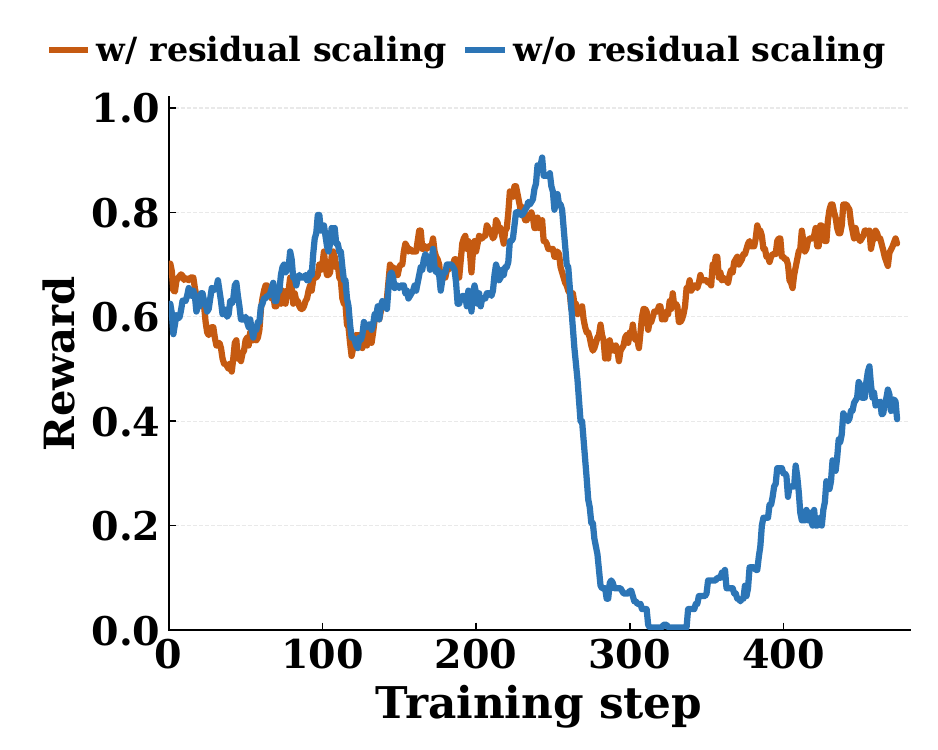}

        \captionof{figure}{  \textbf{Residual scaling preserves stable long-horizon optimization.}
    QwenDLM trained with \textsc{ORL} on GSM8K prompts.
           }
        \label{fig:alpha_schedule_keep}
    \end{minipage}
\end{figure*}

\begin{table}[t]
  \centering
  \footnotesize
  \setlength{\tabcolsep}{3pt}
 \caption{\textbf{Each component matters, and performance is robust to hyperparameters.}
  Accuracy (\%) / TPF on GSM8K and MATH-500. All variants train QwenDLM with
  \textsc{ORL} on GSM8K.
  }
\vspace{-.2cm}
  \label{tab:ablation-sensitivity}
  \begin{tabular}{lcc@{\hspace{10pt}}lcc}
  \toprule
  \multicolumn{3}{c}{\textbf{(a) Training-design ablation}} & \multicolumn{3}{c}{\textbf{(b) Hyperparameter sensitivity}} \\
  \cmidrule(r){1-3}\cmidrule(l){4-6}
  Configuration & GSM8K & MATH-500 & Configuration & GSM8K & MATH-500 \\
  \midrule
  \textbf{Full} & \textbf{94.5 / 6.25} & \textbf{90.0 / 5.64} & \textbf{Default} & \textbf{94.5 / 6.25} & \textbf{90.0 / 5.64} \\
  $-$ AEC & 92.6 / 4.48 & 92.6 / 4.38 & $\lambda_{\mathrm{AEC}}=12.5$ & 94.6 / 5.53 & 92.8 / 5.06 \\
  $-$ entropy-gated RKL & 87.8 / 3.29 & 76.8 / 3.46 & $\lambda_{\mathrm{AEC}}=200$ & 92.6 / 8.53 & 88.2 / 7.42 \\
  $-$ residual pathway & 92.0 / 3.68 & 75.0 / 3.56 & RKL cap $=0.1$ & 93.6 / 6.51 & 92.2 / 5.83 \\
  & & & RKL cap $=20$ & 93.8 / 6.80 & 93.2 / 5.89 \\
  \bottomrule
  \end{tabular}
  \end{table}

\textbf{\textsc{ORL} is highly data-efficient.}
  Figure~\ref{fig:data_perf}(a) compares the amount of additional training
  data used, and
  Figure~\ref{fig:data_perf}(b) the quality--TPF operating point it
  reaches.
  On LLaDA-2.1-mini, \textsc{ORL} consumes 4.2M tokens, the prompts and
  responses of the 5{,}000 rollouts it generates over 625 training steps (one prompt and
  eight rollouts per training step), about $10\times$ less than the roughly 40M tokens of dParallel and
  d3LLM-LLaDA. 
  \textsc{ORL} still reaches a higher
  average TPF at higher average accuracy than every LLaDA-, Dream-, and
  Qwen-based baseline.
  The run takes only  4.5 hours on a single node
  with eight H100 GPUs (36 H100 GPU-hours), and training \textsc{ORL} on QwenDLM
  takes 60 H100 GPU-hours.
  Appendix~\ref{app:training-compute} details how token counts and GPU-hours
  are measured.

\textbf{Residual scaling stabilizes training without sacrificing optimization progress.}
Over the first
250 training steps, verifier reward is nearly identical with and without scaling. The difference emerges over longer training.
With scaling, reward remains high at $0.677$, whereas the unscaled run collapses
to $0.225$ (Figure~\ref{fig:alpha_schedule_keep}).
Removing scaling
changes the backbone update magnitude by only $0.96\times$, but increases
residual-adapter updates by $31.3\times$ (Appendix~\ref{app:residual_scaling}).

\textbf{\textsc{ORL} scales with available refinement capacity.}
Figure~\ref{fig:block_scaling} varies the training block size of
LLaDA-2.1-mini from 32 to 128.
On HumanEval, TPF improves from $11.4$ to $15.0$ at block size 32 and from
$9.2$ to $19.2$ at block size 128.
MBPP shows the same trend (Appendix~\ref{app:block_size_mbpp}).

  \textbf{Effective and robust training.}
  Table~\ref{tab:ablation-sensitivity} reports the contribution of each training
  component and the sensitivity to the main optimization hyperparameters.
  Removing AEC cuts TPF by 28\% on GSM8K and 22\% on MATH-500 while changing
  accuracy by $-1.9$ and $+2.6$ points, so AEC mainly buys speed, and
  $\lambda_{\mathrm{AEC}}$ sets this trade-off (Table~\ref{tab:ablation-sensitivity}(b));
  removing entropy-gated reverse KL or the residual pathway degrades both
  accuracy and TPF, costing up to 15 points on MATH-500.
  AEC also gives a better accuracy--TPF tradeoff than a naive entropy loss at
  different weights (Table~\ref{tab:aec_vs_entropy}).
  In contrast, \textsc{ORL} is robust to its hyperparameters: lowering the AEC
  weight by $4\times$ or varying the RKL cap over a $200\times$ range lowers
  accuracy by at most 0.9 points and mainly shifts TPF.

\noindent
\begin{minipage}[t]{0.60\linewidth}
\vspace{0pt}
\textbf{How on-policy training changes model behavior.}
\label{sec:entropy_dynamics}
Figure~\ref{fig:behavior_stability} (orange) tracks the entropy of the rollout
policy over training: it decreases steadily from $0.20$ to $0.14$ rather than
collapsing, matching the graded increase in TPF.
Figure~\ref{fig:behavior_stability} (blue) tracks the covariance between the
log-probability of sampled tokens and their advantage-weighted probability
(Appendix~\ref{app:theory_entropy_update}),
the quantity that sets the direction of entropy change under a policy-gradient update.
It stays positive on average, so successive updates sharpen the distribution,
and it shrinks toward zero as training proceeds, so sharpening slows down
instead of running away~\citep{cui2025entropy,jin2026revisiting}.
Together with the residual-to-embedding norm ratio, which rises from $0.91$ to
$1.08$ during on-policy training (Figure~\ref{fig:app_residual_norm_ratio}), and the
bounded residual-pathway update ratios in Appendix~\ref{app:residual_scaling},
this indicates that \textsc{ORL} changes how intermediate predictions are
refined across denoising steps rather than merely improving the final-token predictor,
and that it does so by making greater use of residual context without
destabilizing the residual pathway.\par
\end{minipage}\hfill
\begin{minipage}[t]{0.37\linewidth}
\vspace{0pt}
\captionsetup{type=figure}
\centering
\includegraphics[width=\linewidth]{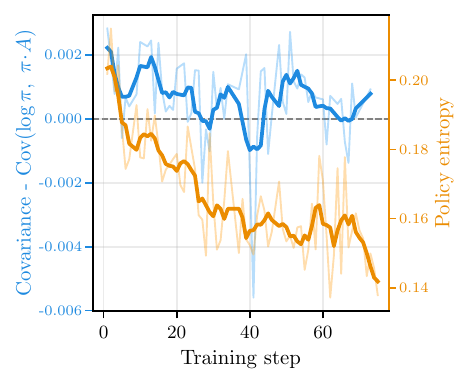}
\caption{Dynamics of on-policy residual learning. The entropy-change covariance
(blue, left axis) remains positive but approaches zero, while rollout
entropy (orange, right axis) decreases gradually without collapse.}
\label{fig:behavior_stability}
\end{minipage}

\noindent
\begin{minipage}[t]{0.53\linewidth}
\vspace{0pt}
\textbf{AEC versus naive entropy minimization.}
Table~\ref{tab:aec_vs_entropy} compares AEC with a naive entropy loss that
uniformly minimizes prediction entropy with weight $\lambda$. A small weight
($\lambda{=}4$) lowers both accuracy and TPF relative to AEC, while a large
weight ($\lambda{=}50$) raises TPF only from 7.7 to 8.2 at a 14.8-point
accuracy cost. Because AEC penalizes only steps whose unresolved entropy fails
to contract, it reaches a better accuracy--TPF tradeoff.
\end{minipage}\hfill
\begin{minipage}[t]{0.44\linewidth}
\vspace{0pt}
\captionsetup{type=table}
\centering
\small
\caption{AEC versus a naive entropy loss on QwenDLM: mean GSM8K accuracy (\%) / TPF
over independent training runs and repeated evaluations.
This ablation trains only on math prompts because of compute limits.}
\label{tab:aec_vs_entropy}
\begin{tabular}{lc}
\toprule
Objective & GSM8K \\
\midrule
AEC ($\lambda_{\mathrm{AEC}}{=}50$) & \textbf{91.6} / 7.7 \\
Naive entropy loss ($\lambda{=}4$) & 90.0 / 7.1 \\
Naive entropy loss ($\lambda{=}50$) & 76.8 / \textbf{8.2} \\
\bottomrule
\end{tabular}
\end{minipage}

\subsection{\texorpdfstring{Deployment Efficiency}{Deployment Efficiency}}
\label{sec:deployment}

\begin{table}[t]
\centering
\caption{Clean-request throughput (tokens/s) of the prior engines and \dlmrt{} for a single request on one H100, with the same checkpoints and decoding settings; red arrows give the speedup of \dlmrt{} over the prior engine. The prior engine is SGLang for QwenDLM and LLaDA-2.1-mini and vLLM for DiffusionGemma.}
\label{tab:engine_comparison}
\footnotesize
\begin{tabular}{@{}llcccc@{}}
\toprule
 & & \multicolumn{2}{c}{GSM8K} & \multicolumn{2}{c}{HumanEval} \\
\cmidrule(lr){3-4}\cmidrule(lr){5-6}
Model & & Prior & \dlmrt{} & Prior & \dlmrt{} \\
\midrule
\multirow{2}{*}{QwenDLM} & Base & 210 & \tpfgain{328}{$1.56\times$} & 164 & \tpfgain{249}{$1.52\times$} \\
 & + \textsc{ORL} & 718 & \tpfgain{987}{$1.37\times$} & 1148 & \tpfgain{1609}{$1.40\times$} \\
\midrule
\multirow{2}{*}{LLaDA-2.1-mini} & Base & 716 & \tpfgain{918}{$1.28\times$} & 1302 & \tpfgain{1698}{$1.30\times$} \\
 & + \textsc{ORL} & 949 & \tpfgain{1281}{$1.35\times$} & 1530 & \tpfgain{2124}{$1.39\times$} \\
\midrule
\multirow{2}{*}{DiffusionGemma} & Base & 990 & \tpfgain{1271}{$1.28\times$} & 963 & \tpfgain{1196}{$1.24\times$} \\
 & + \textsc{ORL} & 1132 & \tpfgain{1456}{$1.29\times$} & 1128 & \tpfgain{1378}{$1.22\times$} \\
\bottomrule
\end{tabular}
\end{table}

\begin{table}[!tb]
\centering
\caption{\textsc{ORL} models in BF16 on one H100 and with NVFP4 quantization on one B200, all throughput measured with \dlmrt{}. Each cell reports accuracy (\%) / TPS; red arrows give the speedup of NVFP4 on B200 over BF16 on H100, which combines the newer GPU with the lower precision. BF16 accuracy is that of Table~\ref{tab:main_quality_tpf} and BF16 TPS that of \dlmrt{} in Table~\ref{tab:engine_comparison}. NVFP4 (NVIDIA) is NVIDIA's NVFP4 checkpoint of the DiffusionGemma base model~\citep{nvidia_diffusiongemma_nvfp4_2026}; NVFP4, TP8 runs DiffusionGemma + \textsc{ORL} with tensor parallelism across eight B200s (generation seed 1234). Appendix~\ref{app:nvfp4_quantization} adds TPF.}
\label{tab:nvfp4_main}
\footnotesize
\setlength{\tabcolsep}{2.5pt}
\begin{tabular}{@{}llcccc@{}}
\toprule
Model & Precision & GSM8K & MATH-500 & HumanEval & MBPP \\
\midrule
\multicolumn{6}{@{}l}{\textit{LLaDA-2.1-mini}} \\
+ \textsc{ORL} & BF16 & 92.0 / 1281 & 87.0 / 1627 & 89.0 / 2124 & 77.3 / 1539 \\
+ \textsc{ORL} & NVFP4 & 92.0 / \tpfgain{2953}{$2.31\times$} & 84.0 / \tpfgain{3882}{$2.39\times$} & 89.0 / \tpfgain{5046}{$2.38\times$} & 75.9 / \tpfgain{3564}{$2.32\times$} \\
\midrule
\multicolumn{6}{@{}l}{\textit{DiffusionGemma}} \\
+ \textsc{ORL} & BF16 & 96.0 / 1456 & 92.3 / 1415 & 91.9 / 1378 & 87.4 / 1361 \\
Base & NVFP4 (NVIDIA) & 96.1 / 3627 & 92.7 / 3198 & 91.3 / 3464 & 87.1 / 2991 \\
+ \textsc{ORL} & NVFP4 & 95.6 / \tpfgain{4214}{$2.89\times$} & 93.4 / \tpfgain{3805}{$2.69\times$} & 91.9 / \tpfgain{3949}{$2.87\times$} & 86.3 / \tpfgain{3898}{$2.86\times$} \\
  + \textsc{ORL} & NVFP4, TP8 & 96.1 / \tpfgain{7708}{$5.29\times$} & 91.4 / \tpfgain{7123}{$5.03\times$} & 92.7 / \tpfgain{7026}{$5.10\times$} & 86.0 / \tpfgain{6871}{$5.05\times$} \\
\bottomrule
\end{tabular}
\end{table}

\begin{table}[!tb]
\centering
\caption{Throughput (tokens/s) of the \textsc{ORL} models with NVFP4 quantization on one consumer GeForce RTX 5090, measured with \dlmrt{} (DiffusionGemma: generation seed 1234).}
\label{tab:rtx5090}
\footnotesize
\ifarxivhighlight\color{arxivred}\fi
\begin{tabular}{@{}lcccc@{}}
\toprule
Model & GSM8K & MATH-500 & HumanEval & MBPP \\
\midrule
LLaDA-2.1-mini + \textsc{ORL} & 1648 & 2081 & 2788 & 2053 \\
DiffusionGemma + \textsc{ORL} & 1540 & 1408 & 1490 & 1465 \\
\bottomrule
\end{tabular}
\end{table}

\textbf{A dedicated engine.}
Table~\ref{tab:engine_comparison} compares \dlmrt{} (Section~\ref{sec:method_inference}) with the prior engines on the same checkpoints and decoding settings, under the measurement protocol of Appendix~\ref{app:eval_details}. Across the measured pairs, \dlmrt{} raises throughput by $1.24$--$1.56\times$ for the Base models and by $1.22$--$1.40\times$ for the \textsc{ORL} models. For QwenDLM and LLaDA-2.1-mini, these measurements use a \dlmrt{} mode that reproduces SGLang's forward pass and decoding decisions ($4{,}816$ of $4{,}820$ QwenDLM answers are identical to SGLang's); for DiffusionGemma, \dlmrt{} follows the reference implementation's BF16 roundings. Across all three families, accuracy under \dlmrt{} stays within $0.6$ points of Table~\ref{tab:main_quality_tpf} (Appendix~\ref{app:dlmrt}).

\textbf{Combining \textsc{ORL} with NVFP4 quantization.}
\textsc{ORL} reduces the number of forward passes, while low-precision inference makes each forward pass cheaper, so the two can be combined. We quantize the routed experts of the \textsc{ORL}-trained LLaDA-2.1-mini and DiffusionGemma to NVFP4, with 4-bit weights and activations, keep attention, routers, and shared experts in BF16, and use an FP8 KV cache. Each 16-weight block takes the FP8 scale that minimizes its layer-output error under a local Hessian of the layer's input activations~\citep{nvidia_local_hessian_2026}; for DiffusionGemma, only the diagonal of this Hessian is used. Because only the scales change, the quantized checkpoints run on standard NVFP4 kernels without extra overhead (Appendix~\ref{app:nvfp4_quantization}).

NVFP4 gives large speedups with small accuracy changes: on one B200, it raises throughput over BF16 on one H100 by $2.31$--$2.39\times$ for LLaDA-2.1-mini and $2.69$--$2.89\times$ for DiffusionGemma (Table~\ref{tab:nvfp4_main}). With \textsc{ORL} and NVFP4, LLaDA-2.1-mini reaches $5{,}046$ tokens/s on HumanEval. DiffusionGemma + \textsc{ORL} in NVFP4 is also $1.14$--$1.30\times$ faster than NVIDIA's NVFP4 release of DiffusionGemma~\citep{nvidia_diffusiongemma_nvfp4_2026} at comparable accuracy (within $0.8$ points; Table~\ref{tab:nvfp4_main}), measured back to back on one B200 with \dlmrt{}. With tensor parallelism across eight B200s, DiffusionGemma + \textsc{ORL} reaches $7{,}708$ tokens/s on GSM8K. Fast decoding does not require datacenter GPUs: the NVFP4 \textsc{ORL} models fit in the 32\,GB of a single consumer GeForce RTX 5090, where \dlmrt{} runs LLaDA-2.1-mini at $2{,}788$ tokens/s on HumanEval and DiffusionGemma at $1{,}540$ tokens/s on GSM8K (Table~\ref{tab:rtx5090}), matching or exceeding BF16 on one H100 on every benchmark (Table~\ref{tab:nvfp4_main}).

\FloatBarrier
\section{Conclusion}
\label{sec:conclusion}
We introduced \textsc{ORL}, a framework for on-policy residual learning in
DLMs.
\textsc{ORL} trains on the verified residual-conditioned
trajectories that the current model induces, adds an entropy-contraction
objective that sharpens predictions across denoising steps, and co-designs the
training system so that exact on-policy supervision is practical.
From QwenDLM, our 4B AR-converted DLM, \textsc{ORL} improves TPF by $2.5$--$6.2\times$ and throughput by approximately $5\times$ at comparable quality (approximately $6.9\times$ together with \dlmrt{}), 
above the measured EAGLE3 and DFlash operating
points.
Applied to LLaDA-2.1-mini and DiffusionGemma, it improves
TPF on all four main benchmarks; on LLaDA-2.1-mini it needs only $4.2$M self-generated tokens. With \dlmrt{} and NVFP4 quantization, LLaDA-2.1-mini + \textsc{ORL} averages $5{,}046$ tokens/s per single request across HumanEval on one B200 and $2{,}788$ tokens/s on a consumer RTX 5090, and DiffusionGemma + \textsc{ORL} averages $4{,}214$ tokens/s across GSM8K on one B200, $1{,}540$ tokens/s on an RTX 5090, and $7{,}708$ tokens/s on eight B200s.

\bibliography{iclr2027_conference}
\bibliographystyle{iclr2027_conference}

\raggedbottom
\appendix
\section{Related Works}
\label{app:related_work}

\paragraph{Residual diffusion language models.}
Retaining continuous information across denoising steps lets a DLM reuse
computation beyond its discrete token commitments, building on
self-conditioning in continuous diffusion~\citep{chen2022analog,jabri2022scalable}.
Loopholing~\citep{jo2026loopholing} introduces a deterministic latent
pathway that bypasses the information loss of categorical sampling;
RCD~\citep{hu2026residual} constructs entropy-weighted residual embeddings
from predictive distributions and trains the model with reference-generated
residual signals; and DiffusionGemma~\citep{team2026diffusiongemma}
incorporates distribution-based self-conditioning into blockwise generation.
Closely related, Relay~\citep{rozonoyer2026learned} trains a cross-step
hidden-state channel through truncated backpropagation through time,
committing ground-truth token values during its training rollouts.
These approaches establish the value of cross-step feedback, but their
training states are typically constructed from ground-truth sequences
or reference-generated signals.
We identify a further challenge: as the model changes, both its residual
predictions and the trajectories they induce change together.
Our multi-seed comparison of every-step and infrequent rollout refresh
shows distinct learning curves (Figure~\ref{fig:motivation}(a)).
To track the evolving distribution, \textsc{ORL} adopts a fully on-policy framework: every update
uses newly generated trajectories from the current model, retaining
verified responses and replaying intermediate states with the exact
residual inputs consumed during generation.
Adjacent entropy contraction encourages faster refinement, while our
co-designed training system makes this workflow practical at scale.

\paragraph{Reinforcement learning and on-policy post-training for DLMs.}
d1~\citep{zhao2026d1} adapts policy-gradient optimization to masked DLMs,
while DiPOD~\citep{jiang2026diffusion} introduces on-policy ELBO regularization
to mitigate drift between its ELBO surrogate and the true log-likelihood.
dUltra~\citep{chen2025dultra} and
LightningRL~\citep{hu2026lightningrl} target decoding efficiency through
learned unmasking decisions and efficiency-aware rewards.
However, generating responses on-policy does not by itself ensure that
training preserves the states encountered during decoding:
LightningRL's released implementation constructs block-masked training
inputs from completed responses rather than replaying the intermediate
decoding states.
TraceRL~\citep{wang2026revolutionizing} aligns training with the decoding
trace, preserving step-level conditioning when no trajectory shrinkage
is applied, but its policy objective supervises the token actions committed
at each step rather than predictions at positions that remain unresolved.
Related non-RL approaches also exploit model-generated trajectories:
d-OPSD~\citep{luo2026learning} constructs a self-teacher by selectively
revealing tokens from generated completions,
dOPSD~\citep{dat2026dopsd} distills later-state predictions into earlier
states, and OPTD~\citep{lu2026optd} learns compressed transitions on
student-visited states.
\textsc{ORL} combines exact residual-state replay with supervision beyond
the sampled actions.
At selected states of verified rollouts, it supervises intermediate
predictions toward the final response, including at positions not committed
at that step.
Adjacent entropy contraction further encourages uncertainty reduction
over the unresolved canvas between consecutive steps, providing a
differentiable signal for denoising progress rather than relying solely
on trajectory-level efficiency rewards.
Both objectives operate within a fully on-policy framework that
regenerates trajectories after every update and preserves the exact
residual context consumed at each training state.

\paragraph{Looped Transformers and latent computation.}
Recurrent computation offers a way to increase reasoning depth without
proportionally increasing model parameters.
Universal Transformers~\citep{dehghani2018universal} introduce shared
transformations with adaptive computation, and subsequent work studies
their reasoning benefits~\citep{saunshi2025reasoning} and scales
latent recurrence to language modeling~\citep{geiping2026scaling,zhu2025scaling}.
LoopFormer~\citep{jeddi2026loopformer} further uses shortcut-consistency
training to support variable computation budgets.
This perspective connects naturally to diffusion generation:
\citet{geiping2025efficient} connect recurrent-depth models to
diffusion through parallel sampling, while
LoopMDM~\citep{lee2026looped} introduces selective layer recurrence
within masked diffusion models.
Residual DLMs realize recurrent computation through continuous
predictive feedback between denoising steps, without requiring additional
layer repetitions within each forward pass.
\textsc{ORL} is complementary to these architectural advances: it trains the
cross-step feedback on fresh trajectories from the current model, with the
exact residual context encountered during generation, and combines
verified-output supervision with an explicit refinement objective so that
intermediate predictions support accurate outputs with fewer decoding steps.

\section{Residual Diagnostics}
\subsection{Predictive Information and Capture Cost}
\label{app:residual_learning_diagnostics}
\label{app:motivation_system_scaling}
\begin{figure}[!htbp]
\centering
\begin{subfigure}[t]{0.48\linewidth}
\centering
\includegraphics[width=\linewidth]{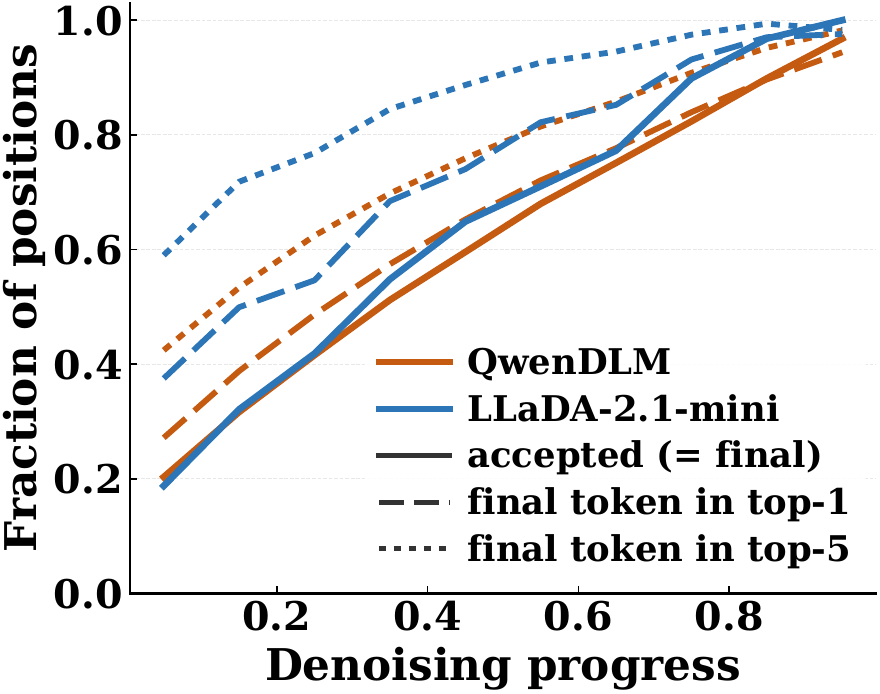}
\caption{Intermediate top-$k$ agreement with final tokens.}
\label{fig:app_topk_agreement}
\end{subfigure}\hfill
\begin{subfigure}[t]{0.48\linewidth}
\centering
\includegraphics[width=\linewidth]{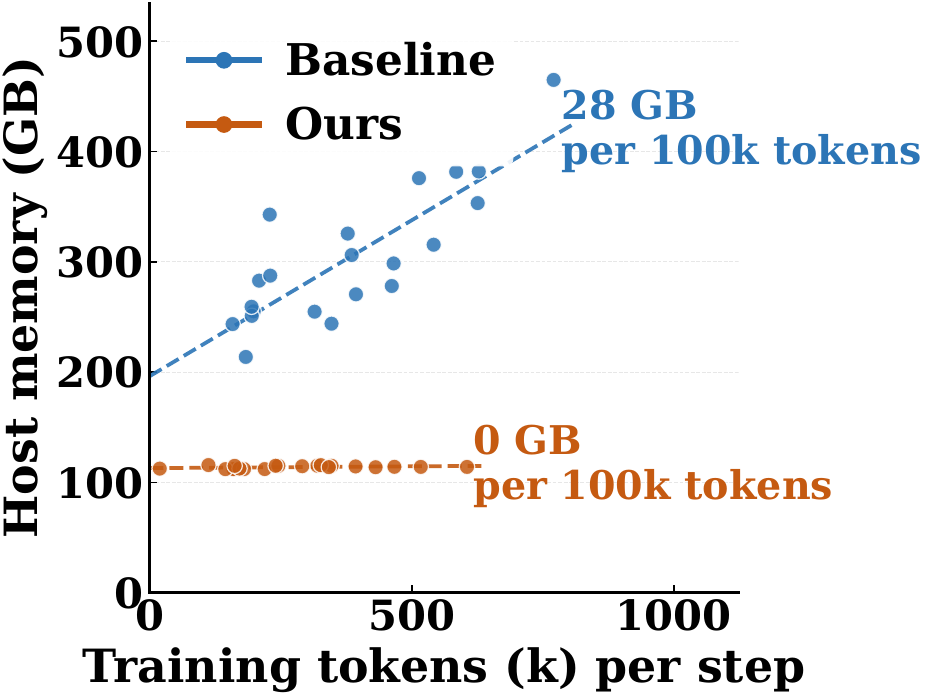}
\caption{Host memory versus materialized training tokens.}
\label{fig:app_motivation_memory}
\end{subfigure}
\caption{Residual-learning motivation. Left: top-$1$/top-$5$ agreement exceeds
the accepted-and-final fraction early in denoising. Right: sparse,
GPU-resident capture avoids the baseline's growing host footprint over
the measured range.}
\end{figure}

Intermediate predictions often contain the final decoded token before it is
committed (Figure~\ref{fig:app_topk_agreement}). This is agreement with the
trajectory's output, not reference-answer correctness. Retaining such states
also creates a memory bottleneck: naive capture adds about 28\,GB of host
memory per 100k training tokens, versus 36\,GB for the combined host/GPU
footprint in Figure~\ref{fig:motivation}(c).

\section{Optimization Details and Stabilization}
\label{app:optimization_stabilization}

\paragraph{Response normalization.}
For verified responses $\mathcal R_{\mathcal B}^{+}$, retained states
$\mathcal S(\mathcal R)$, and valid supervised positions $\mathcal I_b^{\mathcal R}$
(excluding padding and positions after terminal EOS),
let $N_{\mathcal R}=\sum_{(b,k)\in\mathcal S(\mathcal R)}
|\mathcal I_b^{\mathcal R}|$. QwenDLM and LLaDA use
\begin{equation}
 \mathcal A[\ell]=\frac{1}{|\mathcal R_{\mathcal B}^{+}|}
 \sum_{\mathcal R\in\mathcal R_{\mathcal B}^{+}}\frac{1}{N_{\mathcal R}}
 \sum_{(b,k)\in\mathcal S(\mathcal R)}
 \sum_{i\in\mathcal I_b^{\mathcal R}}\ell_{\mathcal R,b,k,i}.
 \label{eq:trajectory_aggregation}
\end{equation}
Thus each verified response has equal token-loss weight; empty verified
batches are skipped.

\paragraph{AEC aggregation.}
A valid pair $(b,k)\in\mathcal P(\mathcal R)$ has a selected target state
$(b,k+1)$ and nonempty unresolved sets at both states. The source need not
be selected: its captured mean normalized entropy supplies the reference.
Give each pair weight $w_{\mathcal R}=1/|\mathcal P(\mathcal R)|$ before
gating. QwenDLM uses $g_{\mathcal R,b,k}=
\mathbf1\{H^{\mathrm{roll}}_{b,k}\le0.20\}$; LLaDA uses $g=1$.
With $D=\sum_{\mathcal R,(b,k)\in\mathcal P(\mathcal R)}
w_{\mathcal R}g_{\mathcal R,b,k}$,
\begin{equation}
 \mathcal L_{\mathrm{AEC}}=
 \frac{\sum_{\mathcal R,(b,k)\in\mathcal P(\mathcal R)}
 w_{\mathcal R}g_{\mathcal R,b,k}\ell^{\mathrm{AEC}}_{\mathcal R,b,k}}{D},
 \qquad D>0,
 \label{eq:aec}
\end{equation}
and the loss is zero for $D=0$. Responses with no valid pair are excluded.
Weights are not recomputed after gating: a response's remaining mass is
its retained fraction of pairs. DiffusionGemma instead uses the canvas-pair
mean in Appendix~\ref{app:efficiency-model-training}.

\paragraph{Entropy-gated supervision.}
For detached reference $q_i^{\mathrm{cap}}$, replayed distribution
$\widehat p_i$, normalized entropy $\overline H$, and cap $C$,
\begin{equation}
 \ell_i^{\mathrm{hybrid}}=
 \begin{cases}
 -\log\widehat p_i(y_i),&\overline H(q_i^{\mathrm{cap}})>\zeta,\\
 \min\{D_{\mathrm{KL}}(\widehat p_i\Vert q_i^{\mathrm{cap}}),C\},
 &\text{otherwise}.
 \end{cases}
 \label{eq:entropy_gated_rkl}
\end{equation}
Only QwenDLM's final training stage uses this gate, with $\zeta=0.10$
and $C=5$; all other reported training stages use CE. The teacher is reconstructed
from each state's captured, shifted output-head-input hidden vector using
the \emph{current} output head under stop-gradient, not an old head or one
final-response teacher. Both head outputs are cast to float32, divided by
loss temperature 1, clipped to $[-50,50]$, then normalized over the vocabulary.
Replay gradients accumulate before one optimizer step.

For $p=\operatorname{softmax}(a)$ and $D_{\mathrm{KL}}(p\Vert q)=D$,
$\partial D/\partial a_j=p_j[\log(p_j/q_j)-D]$ before capping.
Identical transformed distributions therefore give zero RKL loss and gradient.
Exact captured inputs do not guarantee rollout/replay numerical parity;
any nonzero RKL contribution in this training recipe was not measured
separately from the effect of removing low-entropy CE.

\paragraph{Residual initialization.}
QwenDLM's bias-free adapter is
$g_\phi(z)=W_d[\operatorname{SiLU}(W_g u)\odot W_u u]$,
$u=\operatorname{RMSNorm}_{w}(z)$, with hidden/intermediate dimensions
2,560/9,728 and epsilon $10^{-6}$. Multiplying all three matrices by
$\gamma=8$ and dividing both the normalized input and branch output by
$\gamma$ preserves its function:
\begin{equation}
 \widetilde g(z)=\frac{\gamma W_d}{\gamma}
 [\operatorname{SiLU}(\gamma W_g u/\gamma)\odot(\gamma W_u u/\gamma)]
 =g_\phi(z).
 \label{eq:residual_rescaling}
\end{equation}
The RMSNorm gain is unchanged. Both compensation constants are $1/8$,
distinct from injection strength $\alpha=1$; the residual sum then receives
non-affine RMS normalization scaled by 0.02. The same post-normalization
is applied during replay to $z$ in Equation~\ref{eq:residual_input}.
Compensation is applied once,
before initializing the ORL optimizer, not again on resume. Output-only
compensation would not preserve SiLU's arguments.

LLaDA uses affine pre-RMSNorm and three bias-free gated-MLP projections;
activation, dimensions, and normalization epsilon follow the model config.
It initializes gate/up weights from a Gaussian with standard deviation
$10^{-3}$, down weights to zero, and pre-RMSNorm gain to one.
All adapter parameters use $0.3$ times the backbone learning rate.
For LLaDA, self-conditioning probability and
injection strength are both fixed at one; no injection ramp or
post-normalization is used.
Initialization is checked after meta-materialization and before FSDP
synchronization. DiffusionGemma retains its native self-conditioning module.

\section{System Implementation Details}
\label{app:system_details}
Figure~\ref{fig:system_overview} summarizes the \textsc{ORL} loop and where each
mechanism below acts.

\subsection{Asynchronous Exact-State Capture}
\label{app:async_capture}
Each forward snapshots mutable token IDs and accept/edit masks, retains the
residual and required GPU statistics, records a CUDA event, and enqueues a
bounded handle. A background worker materializes the record after the event;
the decode critical path performs no host copy or host synchronization.
Exhausting the queue's entry/byte budget blocks for a bounded interval or
fails explicitly, never silently dropping states. Host-bound fields use
pinned memory. With a colocated trainer, residual/hidden-source pages can
remain GPU-resident and cross through one-shot CUDA-IPC handles, relaying
only required pages to another GPU. Storage remains live until consumer
acknowledgement.

\subsection{Staircase Attention}
\label{app:staircase_kernel}
A noisy block attends to the prompt, preceding clean blocks, and itself,
but not its clean target, later blocks, other noisy blocks, or other
documents. Clean queries never attend to noisy keys. The Triton kernel
derives visible key ranges from per-document metadata, skips invisible tiles,
and applies the predicate within boundary tiles; it allocates no dense
$N\times N$ mask. Backward uses the same visibility: $dQ$ follows the
query traversal, while $dK,dV$ derive valid query ranges per key tile.
The implementation supports variable-length packed documents, grouped-query
attention, and sequence parallelism.
Our \href{\orlcodeurl}{code repository} contains the version of this kernel used to train and
reproduce our models.\footnote{The latest version is maintained at
\url{https://github.com/yuezhouhu/fast-diffusion-attention}, which provides kernels derived from
FlashAttention-2 and FlashAttention-3 and matches FlashAttention's numerical accuracy on SM80, SM86,
SM89, and SM90 GPUs.}

\subsection{Packing and Cost Balancing}
\label{app:packed_balanced}
Same-round states share clean context, following the slicing structure of
TraceRL~\citep{wang2026revolutionizing}; residual/source pages remain sparse
and block-local. Replay rows are packed as mutually invisible documents.
A weighted proxy of sequence tokens, supervised positions, and resident
pages drives deterministic largest-cost-first assignment to data-parallel
groups. Chunked materialization bounds the live input/page footprint.
Under hybrid underfill, splittable label sets are partitioned before
zero-loss ghost rows align collective schedules. Each label has one owner,
and global reduction uses $\mathcal L=\sum_j\mathrm{num}_j/
\sum_j\mathrm{den}_j$, preserving the unpacked objective up to rounding.

\section{Training and Evaluation Details}
\label{app:training_details}

\subsection{Evaluation Protocols}
\label{app:eval_details}
Measured throughput uses one H100, TP=1, and a single request in
flight. Length-terminated responses count as incorrect and are
excluded only from clean efficiency statistics. Request time includes
prefill, decoding, scheduling, and host overhead, but excludes model loading,
server startup, and preprocessing. Warmup exclusions are stated per run.
Within each family, Base and ORL share tasks and scorers. Each main-table
evaluation record is associated with its checkpoint and evaluation profile
in the code repository (Appendix~\ref{app:generation_configs}). Runtime
acceleration is evaluated separately in
Appendix~\ref{app:inference-optimization}.

\paragraph{Runtime settings.}
QwenDLM and LLaDA use SGLang~\citep{zheng2024sglang}; DiffusionGemma uses vLLM~\citep{kwon2023efficient} in BF16.
Reproducing the throughput results of Figure~\ref{fig:teaser}(b), Section~\ref{sec:deployment}, and
Figure~\ref{fig:appendix-gsm-he-baselines} requires \dlmrt{} (Section~\ref{sec:method_inference}).
Engine revisions, container digests, attention backends, cache controls,
and graph settings are recorded in \path{evaluation/configs/paper/decoding/}
in the code repository. Absolute TPS is hardware/runtime-specific.
Separate B200 NVFP4 measurements, including the
quantization procedure and FP8 KV-cache configuration,
are reported in Appendix~\ref{app:nvfp4_quantization}.

\paragraph{Tasks and scoring.}
The main suite uses GSM8K test (1,319 tasks)~\citep{cobbe2021training}, MATH-500 (500)~\citep{hendrycks2021measuring}, HumanEval
(164, base tests)~\citep{chen2021evaluating}, and the canonical MBPP427 collection~\citep{austin2021program} with its original
multi-candidate extraction. Base/ORL share task and scorer protocols within
each family; cross-family and literature comparisons are not scorer-matched.
QwenDLM retains its CodeTACO code-scoring path, while LLaDA/DiffusionGemma
use EvalPlus base tests for HumanEval. In the code repository,
\path{evaluation/repro/paper/main.yaml} and
\path{evaluation/repro/eval/portable/protocols.json}
record input identities, native prompting, extraction rules, and scorer pins.

\paragraph{Runs and aggregation.}
For each ORL configuration, we conduct multiple independent training runs
and repeated evaluations of each resulting checkpoint. In the main benchmark
suite, the same checkpoint from each training run is evaluated across all
benchmarks. We first average the evaluation metrics within each checkpoint,
then average equally across independent training runs. This averaging
protocol also applies to the component and hyperparameter ablations in
Table~\ref{tab:ablation-sensitivity}. Starting-model baselines average repeated
evaluations of their reference checkpoints. Each evaluation generates one
completion per task; DiffusionGemma uses three generation-seed settings per
checkpoint. Seed settings and task partitions are recorded in
\path{evaluation/repro/paper/main.yaml}; transfer-evaluation settings are given below.
Reported means alone do not establish statistical significance for small
score differences. Within an individual HumanEval evaluation, one task
corresponds to approximately 0.61 percentage points; averaging repeated
evaluations permits smaller increments in the reported means.

\paragraph{Reproduction configurations.}
\label{app:generation_configs}
For the complete generation configuration associated with each main-table
checkpoint and operating point, refer to our
\href{\orlcodeurl}{code repository}.
The experiment index at \path{evaluation/configs/paper/experiments.yaml} maps results
to their profiles in \path{evaluation/configs/paper/decoding/};
\path{evaluation/docs/reproduction/SETTINGS.md} provides the configuration guide.
Profiles outside the main table are defined where they are reported
(Tables~\ref{tab:spec_context} and~\ref{tab:matched_decoder_grid};
Appendix~\ref{app:quality_throughput_frontiers}).

\paragraph{Accounting.}
Let $\mathcal I_b$ be all requests, $\mathcal C_b$ those not length-terminated,
$s_i$ the scorer outcome, $C_i$ completion tokens, $D_i$ denoising forwards,
and $W_i$ complete request time. We report
\begin{align}
 \operatorname{Acc}_b&=\frac{100}{|\mathcal I_b|}
       \sum_{i\in\mathcal C_b}s_i,
 &\operatorname{TPF}_b&=\frac{1}{|\mathcal C_b|}
       \sum_{i\in\mathcal C_b}\frac{C_i}{D_i},
 \label{eq:eval-tpf}\\
 \operatorname{TPS}^{\mathrm{clean}}_b
 &=\frac{\sum_{\mathcal C_b}C_i}{\sum_{\mathcal C_b}W_i},
 &\operatorname{TPS}^{\mathrm{all}}_b
 &=\frac{\sum_{\mathcal I_b}C_i}{\sum_{\mathcal I_b}W_i},
 \label{eq:eval-wall-tps}\\
 \operatorname{TPS}^{\mathrm{completed}}_b
 &=\frac{\sum_{\mathcal C_b}C_i}{\sum_{\mathcal I_b}W_i}.
 \label{eq:eval-all-request-tps}
\end{align}
Completed-output TPS charges failed-request time without crediting its
tokens, but still credits completed incorrect answers; it is not
correct-answer throughput. All-forward TPF replaces $D_i$ by
$F_i=D_i+R_i+K_i$ (denoise, refinement, commit). Wall TPS includes all this
work; the approximation $\operatorname{TPF}/t_{\mathrm{forward}}$ does not.
We compute benchmark metrics for each evaluation and average repeated
evaluations within each checkpoint. For ORL, we then average equally across
independent training runs. Cross-benchmark summaries take an equal-weight
mean of the benchmark-level metrics.

\paragraph{Baseline sources.}
\label{app:baselines}
The underlined starting points and ORL rows in
Table~\ref{tab:main_quality_tpf} are ours; other rows are paper-reported.
Fast-dLLM--LLaDA~\citep{wu2025fast}, dParallel--LLaDA~\citep{chen2026dparallel},
D2F--LLaDA~\citep{wang2026diffusion}, Fast-dLLM-v2-7B~\citep{wu2026fast}, and d3LLM
follow the unified d3LLM evaluation~\citep{qian2026d3llm}, whose MATH entries are MATH (4-shot) on the full test set rather than MATH-500;
SDAR-8B-b32~\citep{cheng2025sdar} follows LightningRL~\citep{hu2026lightningrl}.
RCD~\citep{hu2026residual}, dUltra-coding~\citep{chen2025dultra}, and
DMax-Math~\citep{chen2026dmax} use their original papers.
TraDo-4B (dynamic) is the TraceRL-trained model reported by
\citet{wang2026revolutionizing}.
For dOPSD~\citep{dat2026dopsd}, we quote the LLaDA-8B-Instruct results
from Table~1: GSM8K/MATH-500 accuracy $72.87/36.00$ and
HumanEval/MBPP pass@1 $39.63/39.54$, with math/code averages
$54.44/39.59$. Corresponding TPF is unreported and marked with dashes.
These published comparisons retain their original backbones and evaluation
settings; they are not scorer-matched reruns.

Derived comparison points are estimates, not measured wall TPS. Their
conversion is $\widehat{\mathrm{TPS}}_{m,b}=c_b\mathrm{TPF}_{m,b}$, with
$c_b$ taken from a measured reference's TPS/TPF or the mean of two matched
endpoint ratios. Disaggregating a paper's aggregate TPS instead uses
$\widehat{\mathrm{TPS}}_{m,b}=\mathrm{TPS}^{\mathrm{paper}}_m
\mathrm{TPF}_{m,b}/\overline{\mathrm{TPF}}_m$ under that paper's task,
hardware, and decoding protocol.

\subsection{Additional Benchmarks}
\label{app:full_evaluation}
Table~\ref{tab:transfer_eval_protocol} summarizes the transfer suite:
AIME-24~\citep{aime}, GPQA-Diamond~\citep{rein2024gpqa}, MMLU-Pro~\citep{wang2024mmlu},
LiveCodeBench-v6~\citep{jain2024livecodebench}, and IFEval~\citep{zhou2023instruction}.
In the code repository, \path{evaluation/repro/paper/transfer.yaml} and
\path{evaluation/repro/paper/TRANSFER_GENERATION.md} record exact seeds, input hashes,
prompt serialization, context settings, and scorer profiles.

\begin{table}[t]
\centering
\small
\setlength{\tabcolsep}{6pt}
\caption{Transfer evaluation. Evaluation-seed counts are generation-seed settings
per trained checkpoint (LL: LLaDA; DG: DiffusionGemma).
ORL metrics average independent training runs and repeated evaluations;
starting-model baselines average repeated evaluations of their reference
checkpoints. Each evaluation generates one completion per task.
AIME-24 scores 30 tasks from a shared 60-task, two-year generation input.}
\label{tab:transfer_eval_protocol}
\begin{tabular}{lrr}
\toprule
Benchmark & Tasks & Eval. seeds LL/DG \\
\midrule
AIME-24 & 30 & 8 / 8 \\
GPQA-Diamond & 198 & 3 / 3 \\
MMLU-Pro & 12,032 & 1 / 3 \\
LiveCodeBench-v6 & 175 & 1 / 3 \\
IFEval & 541 & 1 / 3 \\
\bottomrule
\end{tabular}
\end{table}

Quality and clean per-response TPF are first averaged over repeated
evaluations of each checkpoint; ORL metrics are then averaged equally across
independent training runs. Quality is not best-of-$k$ or majority voting.
AIME uses numeric parsing,
GPQA/MMLU-Pro the original multiple-choice parser, LiveCodeBench execution
tests, and IFEval its instruction evaluator. Historical source/RNG binding
for the LCB/IFEval reproduction scorers and the DiffusionGemma long-task
context override remains incomplete; the repository records these separately
from fresh-run settings.

\paragraph{Transfer quality.}
In Table~\ref{tab:cross_domain_generalization}, TPF increases on all ten
model--benchmark pairs, while accuracy changes vary by task. LLaDA's accuracy
declines by 3.4, 1.8, and 1.7 percentage points on GPQA-Diamond, IFEval, and
AIME-24, respectively; DiffusionGemma's accuracy declines by 2.1 percentage
points on LiveCodeBench-v6.

\subsection{Offline SDAR Initialization}
\label{app:offline_sdar_training}
QwenDLM starts from a non-residual Qwen3-4B-Thinking-2507 SDAR checkpoint.
We add a zero-output residual adapter and train with CE on
ReasonMix/SFTMix, using block size 256 and an 8,192-token cutoff.
The initialization checkpoint for the on-policy experiments and the full
offline recipe, including optimizer, normalization, and hardware settings, are
recorded in \path{evaluation/configs/paper/training.yaml}; the runnable
offline recipes are in \path{offline/recipes/}.
An auxiliary sweep evaluated 118 checkpoints on GSM8K100 and HumanEval164;
later checkpoints did not monotonically improve accuracy.

\subsection{On-Policy Training Recipes}
\label{app:efficiency-model-training}
We train each model until tokens per forward (TPF) stabilizes.
Each iteration draws one prompt and eight rollouts and updates on verified
responses only; batches with no verified response do not trigger an optimizer
update. \path{evaluation/configs/paper/training.yaml} in the code repository
records the complete per-model recipes, training schedules, checkpoint identities,
rollout controls, adapter initialization, optimizer settings, resume records, and
compute accounting; the runnable recipes are in \path{training/recipes/}.

\paragraph{Data.}
QwenDLM uses 819 OpenCodeInstruct-derived Python problems. LLaDA and
DiffusionGemma use a balanced mixture of those code problems, 819 GSM8K
problems, and 819 MATH problems. Code rewards use the supplied test suite;
math rewards use task-specific verifiers. Exact normalized checks found
no final-training-file matches against the corresponding HumanEval,
GSM8K, and MATH-500 evaluation sets; semantic and earlier-stage contamination
are not established by this check.

\paragraph{QwenDLM.}
After residual rescaling by $\gamma=8$, we train at learning rate $10^{-5}$,
then continue at $5\times10^{-6}$ with the
entropy-gated CE/RKL loss. Response-normalized CE is combined with AEC
(coefficient 50, $\rho=0.05$, source-entropy ceiling 0.20), using the state
selection in Section~\ref{sec:method}.

\paragraph{LLaDA-2.1-mini.}
We add the zero-output adapter and train with block size 32, backbone/adapter
learning rates $10^{-5}/3\times10^{-6}$, and response-normalized CE plus AEC
(coefficient 50, $\rho=0.20$), without entropy-gated RKL.
Length-truncated responses are rejected.

\paragraph{DiffusionGemma.}
We retain native self-conditioning and train with a 256-token canvas,
keeping every intermediate state. Gradients pass through the current canvas
forward; captured residuals and prefix-cache construction are detached.
Token-mean CE is combined with the mean over valid adjacent canvas pairs
within the same block:
\begin{equation}
 \mathcal L_{\mathrm{AEC}}^{\mathrm{DG}}=
 \frac{1}{|\mathcal P_{\mathrm{DG}}|}
 \sum_{(t,t+1)\in\mathcal P_{\mathrm{DG}}}
 [H^\theta_{t+1}-0.8\,\operatorname{sg}(H^\theta_t)]_+^2 .
 \label{eq:dg_aec}
\end{equation}
Both entropies are recomputed by the actor, and the source is detached.
The AEC coefficient is 100, with no entropy gate. AdamW uses a learning rate
$10^{-5}$. Training uses an eight-GPU actor and a separate eight-GPU rollout worker.

\section{Inference Optimization}
\label{app:inference-optimization}
\label{app:commit-forward-folding}
We reduce inference overhead through cache reuse, fused execution, and CUDA
graphs (Section~\ref{sec:method_inference} gives an overview). Implementation details and configuration-specific compatibility
checks are provided in the code repository
(Appendix~\ref{app:generation_configs}). The final output-parity gate covers
HE164, EvalPlus MBPP378, and GSM8K100.

\begin{figure}[!htbp]
\centering
\includegraphics[width=0.94\linewidth]{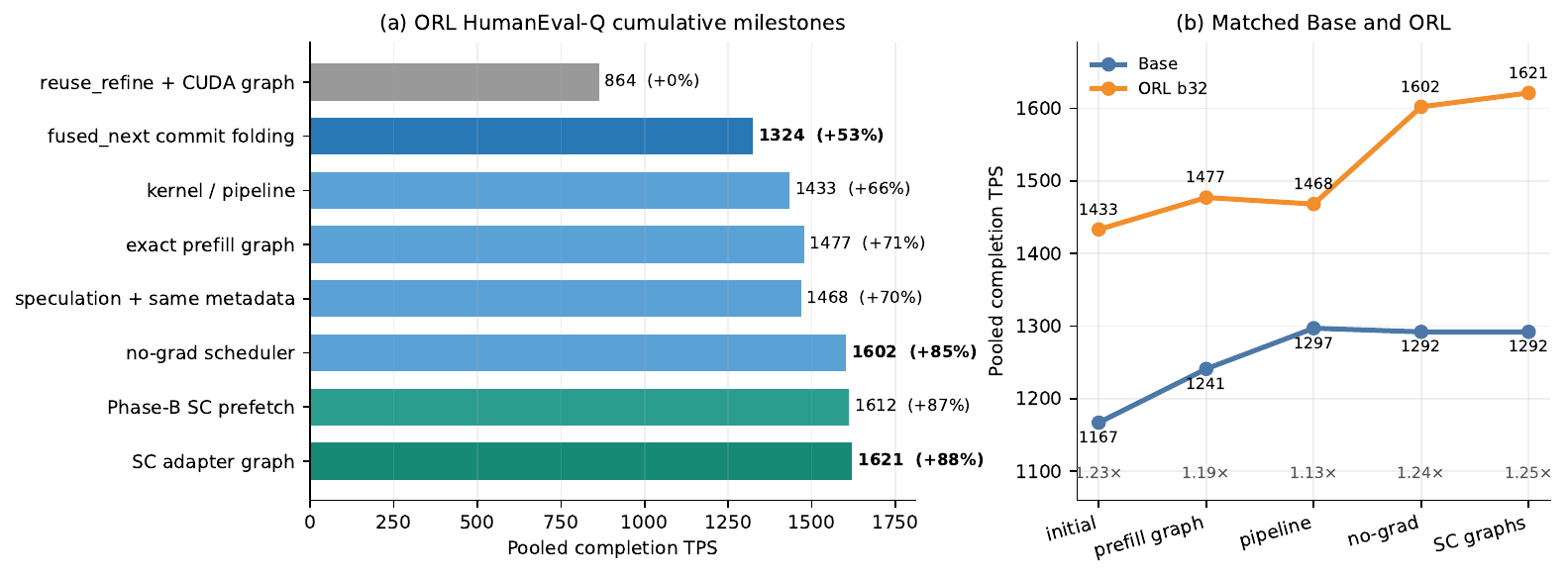}
\caption{Cumulative LLaDA inference optimizations in SGLang, HumanEval Quality mode.
Each side uses four one-H100 engines, pooled completion TPS, and removes
the first warmup request per shard. The ORL checkpoint differs
from the main-table checkpoint. The 864-to-1,324 transition had 153/164
identical outputs and accuracy $0.896\to0.890$, not exact parity.
Later changes pass the final gate. The last two normalized gains are
0.60\%/1.18\%; their raw paired comparison is $1610.2\to1629.2$ TPS.}
\label{fig:llada-inference-tps-cumulative}
\end{figure}

\paragraph{Cost and scope.}
\label{app:tpf_tps_gap}
Before the later graph fixes, ORL improves TPF by $1.51\times$ but TPS by
$1.23\times$: Base/ORL forwards take 7.47/9.17\,ms. After optimization,
they take 6.74/8.11\,ms. Self-conditioning adds work and fewer forwards
amortize fixed costs less well, so TPF is not a throughput substitute.
Supported execution modes and compatibility requirements are specified in
the repository. Output parity is rechecked on new hardware.

\subsection{\texorpdfstring{The \dlmrt{} Engine}{The dlmrt Engine}}
\label{app:dlmrt}

\dlmrt{} serves one request at a time. A block of 32 tokens (LLaDA-2.1-mini) or 256 tokens (QwenDLM and DiffusionGemma) starts as noise and is refined by denoising rounds until the decoder accepts every position; a causal commit pass then writes the block's keys and values for the blocks after it.

\textbf{The block graph.} \dlmrt{} captures each block as one CUDA graph (Figure~\ref{fig:dlmrt_block_graph}). The denoising loop is a CUDA conditional \texttt{WHILE} node whose condition kernel reads device-side counters after every round, so no round waits for the host: the host launches one graph per block and reads back the block's token IDs. For LLaDA-2.1-mini, a finished block's refinement rounds run as the first half of the next block's first forward pass, which is the commit folding of Section~\ref{sec:method_inference}; DiffusionGemma commits each canvas at the start of the next canvas's graph. \dlmrt{} also runs tensor-parallel across an NVLink node with a barrier-free fused reduction; except for the TP8 row of Table~\ref{tab:nvfp4_main}, all measurements in this paper use one GPU.

\begin{figure}[!htbp]
\centering
\begin{minipage}{0.9\linewidth}
\ifarxivhighlight\color{arxivred}\fi
\small
\textbf{Host, once per block:} launch the block's CUDA graph; read back its token IDs; check EOS and the token budget.\\[3pt]
\textbf{Device, one CUDA graph per block:}
\begin{algorithmic}[1]
\State $\mathbf{x} \gets$ prompt tail or noise
\While{some position is not accepted \textbf{and} round $<$ limit} \Comment{conditional node}
\State self-conditioning from the previous round's probabilities; embedding
\State $L\times$[norm, QKV, Q/K norm and RoPE, KV write, attention, output projection, norm, MLP or MoE]
\State final norm, LM head, row statistics; accept, edit, or re-noise
\EndWhile
\State commit: causal forward pass over the accepted block, writing its keys and values
\end{algorithmic}
\end{minipage}
\caption{One block in \dlmrt{}. The denoising loop runs on the GPU inside one CUDA graph; the host only launches the block and reads back its tokens.}
\label{fig:dlmrt_block_graph}
\end{figure}

\textbf{Kernels.} Every kernel in a round is a programmatically dependent launch: it waits for its predecessor only before touching memory, so its launch overlaps the predecessor's tail (about $0.4\,\mu$s saved per kernel boundary). At 256 rows, the projections sit near the H100's compute--bandwidth ridge (Table~\ref{tab:dlmrt_gemm}). \dlmrt{} runs them on DeepGEMM on Hopper, patched with an L2 policy that streams weights and keeps activations resident ($1$--$1.5\%$ faster per round), and on cuBLAS on Blackwell. On Hopper it fuses gate\_up and SwiGLU into one ThunderKittens kernel ($3$--$4\%$ per round) and, for QwenDLM, the output projection, residual RMSNorm, gate\_up, SwiGLU, and down projection into one persistent kernel ($1.6$--$2.3\%$ per round); all three changes are bit-identical to the unpatched, unfused kernels. Attention uses FA3 over one contiguous KV arena with device-side sequence lengths, and a Triton flash attention on Blackwell. Q/K normalization, RoPE, and the KV write run as one fused row kernel, and the self-conditioning softmax reuses the decoder's log-normalizer, which cuts it from $412$ to $50\,\mu$s. MoE routing runs in Triton; the experts use DeepGEMM's masked grouped GEMM in BF16, small-tile Triton grouped GEMMs for LLaDA-2.1-mini's 32-row rounds, and FlashInfer's TRT-LLM-gen NVFP4 MoE on Blackwell (Appendix~\ref{app:nvfp4_quantization}).

\begin{table}[!htbp]
\centering
\small
\ifarxivhighlight\color{arxivred}\fi
\caption{Time per call ($\mu$s) of the QwenDLM-4B projections at 256 rows on one H100, against the time to read the BF16 weights at peak HBM bandwidth.}
\label{tab:dlmrt_gemm}
\begin{tabular}{lrrr}
\toprule
Projection & cuBLAS BF16 & DeepGEMM BF16 & HBM floor \\
\midrule
QKV ($2560\to6144$) & 15.7 & 15.6 & 9.4 \\
O ($4096\to2560$) & 13.6 & 12.1 & 6.3 \\
gate\_up ($2560\to19456$) & 46.2 & 45.1 & 29.7 \\
down ($9728\to2560$) & 30.8 & 26.8 & 14.9 \\
\bottomrule
\end{tabular}
\end{table}

\textbf{Numerics.} \dlmrt{} keeps each model's decoding rule. For QwenDLM and LLaDA-2.1-mini, an SGLang-numerics mode runs SGLang's forward pass and decisions inside \dlmrt{}'s block graphs, with kernels that reproduce SGLang's reductions bit for bit where they are faster; Table~\ref{tab:engine_comparison} uses this mode. Against SGLang's records, it reproduces $4{,}816$ of the $4{,}820$ answers of the two QwenDLM models on the four main benchmarks, and all 164 HumanEval answers of LLaDA-2.1-mini with \textsc{ORL} in Speed mode. For DiffusionGemma, every kernel follows the reference implementation's BF16 roundings, and a parity test checks the forward pass against it. Across all three families, accuracy under \dlmrt{} stays within $0.6$ points of Table~\ref{tab:main_quality_tpf}.

\textbf{Latency.} Table~\ref{tab:dlmrt_round} lists the latency of one denoising round, from self-conditioning to the decoder's decision, at an empty context.

\begin{table}[!htbp]
\centering
\small
\ifarxivhighlight\color{arxivred}\fi
\caption{Latency of one denoising round in \dlmrt{} at an empty context on one GPU (median over replays of one captured round).}
\label{tab:dlmrt_round}
\begin{tabular}{llr}
\toprule
Model & GPU, precision & Round (ms) \\
\midrule
QwenDLM + \textsc{ORL} & H100, BF16 & 5.04 \\
LLaDA-2.1-mini + \textsc{ORL} & H100, BF16 & 4.99 \\
 & B200, NVFP4 experts & 2.03 \\
DiffusionGemma + \textsc{ORL} & H100, BF16 & 16.96 \\
 & B200, NVFP4 experts & 5.59 \\
\bottomrule
\end{tabular}
\end{table}

\subsection{NVFP4 Quantization}
\label{app:nvfp4_quantization}

To improve inference throughput and latency on Blackwell GPUs, we
quantize the ORL models to NVFP4 with 4-bit weights and activations
(W4A4). The main challenge is preserving generation quality at this
precision. We use NVIDIA's Local-Hessian scale search
algorithm~\citep{nvidia_local_hessian_2026} to select weight scales
that minimize layer-output reconstruction error. For mixture-of-experts
(MoE) models, we quantize only the routed experts; attention, routers,
shared experts, and all other components remain in their original
BF16 precision. Our NVFP4 inference configurations also use an FP8
KV cache. The NVFP4 measurements below use B200 GPUs;
the H100 rows repeat the single-H100 BF16 results of the rest of the paper.
Section~\ref{sec:deployment} summarizes the recipe and results.

\paragraph{Local-Hessian scale search.}
Consider a linear layer $Y=WX$, where $W$ is the weight matrix and
$X$ is the input activations. For reconstructed quantized
weights $\hat W$, the squared norm of the layer-output reconstruction error is
\begin{equation}
\mathcal E
= \|(\hat W-W)X\|_F^2
= \operatorname{tr}\!\left(
    (\hat W-W)H(\hat W-W)^\top
  \right),
\qquad H=XX^\top.
\label{eq:nvfp4_output_error}
\end{equation}
The input second-moment matrix $H$ is proportional to the Hessian of
this reconstruction objective. Unlike weight mean-squared error,
this objective accounts for both the importance of individual input
features and their correlations.

Local-Hessian makes the search tractable by approximating $H$ as
block diagonal and optimizing each 16-weight NVFP4 block independently.
For a weight block $w_b$, it selects
\begin{equation}
s_b^*
= \arg\min_{s\in\mathcal S}
  \bigl(\hat w_b(s)-w_b\bigr)^\top
  H_b
  \bigl(\hat w_b(s)-w_b\bigr),
\qquad
H_b=X_bX_b^\top\in\mathbb R^{16\times16},
\label{eq:nvfp4_scale_search}
\end{equation}
where $X_b$ contains the corresponding input features and
$\hat w_b(s)=s\,Q_{\mathrm{FP4}}(w_b/s)$ denotes the reconstructed
weights. The search evaluates all 126 positive, nonzero FP8 E4M3
scale candidates and chooses the scale with the smallest local output
error. This approximation retains correlations within each block
while ignoring interactions across blocks. Because the procedure
changes only the quantization scales, the resulting weights remain
compatible with standard NVFP4 inference kernels and introduce no
additional runtime overhead.

\paragraph{Diagonal-Hessian variant.}
We also evaluate a simpler variant that replaces $H_b$ in
Equation~\ref{eq:nvfp4_scale_search} with its diagonal matrix.
This variant weights reconstruction errors by the second moment of
each input feature but ignores all off-diagonal correlations, including
those within a 16-weight block.
We test both variants and select the quantized checkpoint with the
best generation quality. The selected variants are Local-Hessian
for LLaDA-2.1-mini-ORL and Diagonal-Hessian for DiffusionGemma-ORL.

\begin{table}[t]
\centering
\small
\setlength{\tabcolsep}{4pt}
\resizebox{\linewidth}{!}{%
\begin{tabular}{llcccc}
\toprule
Model & Configuration & GSM8K & MATH-500 & HumanEval & MBPP \\
\midrule
LLaDA-2.1-mini-ORL
 & BF16 (H100)
 & 92.0 / 8.5 / 1281
 & 87.0 / 10.2 / 1627
 & 89.0 / 13.1 / 2124
 & 77.3 / 10.3 / 1539 \\
 & NVFP4 (Local-Hessian, B200)
 & 92.0 / 7.54 / 2953
 & 84.0 / 9.72 / 3882
 & 89.0 / 12.23 / 5046
 & 75.9 / 9.79 / 3564 \\
\midrule
DiffusionGemma-ORL
 & BF16 (H100)
 & 96.0 / 30.0 / 1456
 & 92.3 / 27.4 / 1415
 & 91.9 / 27.9 / 1378
 & 87.4 / 28.7 / 1361 \\
 & NVFP4 (Diagonal-Hessian, B200)
 & 95.6 / 29.25 / 4214
 & 93.4 / 26.69 / 3805
 & 91.9 / 27.32 / 3949
 & 86.3 / 27.96 / 3898 \\
\midrule
DiffusionGemma
 & NVFP4 (NVIDIA, B200)
 & 96.1 / 24.81 / 3627
 & 92.7 / 22.79 / 3198
 & 91.3 / 23.91 / 3464
 & 87.1 / 21.60 / 2991 \\
\bottomrule
\end{tabular}%
}
\caption{BF16 and NVFP4 evaluation of ORL models.
Each cell reports accuracy (\%) / tokens per forward (TPF) /
tokens per second (TPS). Each configuration runs on one GPU of the
indicated type, with throughput measured with \dlmrt{}. H100 rows list the single-H100 BF16 results:
accuracy and TPF as in Table~\ref{tab:main_quality_tpf}, TPS as in Table~\ref{tab:engine_comparison}.
B200 rows report NVFP4 measurements; DiffusionGemma results are averaged over generation seeds
1234--1236, with the ORL decoder on all four tasks. NVFP4
configurations use an FP8 KV cache. The throughput gains in Section~\ref{sec:deployment} compare
NVFP4 on B200 with BF16 on H100.}
\label{tab:nvfp4_results}
\end{table}

Table~\ref{tab:nvfp4_results} shows that NVFP4 on B200 increases
throughput by $2.31$--$2.39\times$ for LLaDA-2.1-mini-ORL and
$2.69$--$2.89\times$ for DiffusionGemma-ORL relative to the H100 BF16
results. Accuracy remains broadly comparable, although
LLaDA loses 3.0 percentage points on MATH-500 and 1.4 points on MBPP
against the H100 results.
TPF decreases modestly across all four tasks for both models.

\section{Throughput and Additional Comparisons}
\label{app:additional_compairson}
\subsection{Full-Task Request Accounting}
\label{app:throughput_accounting}
Table~\ref{tab:throughput_accounting} covers the same GSM1319/HE164 IDs on
both sides, with no warmup exclusion. Qwen problem text, gold, and prompt-token
counts match; wire prompts were not archived. LLaDA lacks prompt text and HE
gold tests, so task-ID agreement alone does not establish prompt/test parity.
\begin{table}[t]
\centering
\footnotesize
\setlength{\tabcolsep}{3.5pt}
\caption{Full-task main runs. $L/N$ counts length failures over all
tasks. Lengths are mean returned tokens on clean/all requests. TPS columns
use Equations~\ref{eq:eval-wall-tps} and~\ref{eq:eval-all-request-tps}:
clean, all-output, and completed-output/all-request-time, respectively.
Quality always includes all tasks, with length failures incorrect.
All runs use SGLang.}
\label{tab:throughput_accounting}
\begin{tabular}{llrrrrrr}
\toprule
Role & Task & Acc. (\%) & $L/N$ & Length clean/all & Clean TPS & All TPS & Completed TPS \\
\midrule
\multicolumn{8}{l}{\emph{QwenDLM}} \\
Base & GSM8K & 90.30 & 1/1319 & 1723/1759 & 209.7 & 213.5 & 209.0 \\
ORL & GSM8K & 92.42 & 1/1319 & 1140/1164 & 718.4 & 724.8 & 709.3 \\
Base & HumanEval & 82.93 & 6/164 & 4929/5948 & 164.1 & 196.0 & 156.5 \\
ORL & HumanEval & 89.63 & 4/164 & 5621/6283 & 1148.2 & 1153.7 & 1007.0 \\
\midrule
\multicolumn{8}{l}{\emph{LLaDA-2.1-mini}} \\
Base & GSM8K & 91.89 & 11/1319 & 265/297 & 715.7 & 669.1 & 592.1 \\
ORL & GSM8K & 92.04 & 2/1319 & 216/222 & 949.2 & 938.7 & 912.5 \\
Base & HumanEval & 86.59 & 0/164 & 360/360 & 1302.3 & 1302.3 & 1302.3 \\
ORL & HumanEval & 89.02 & 0/164 & 546/546 & 1530.1 & 1530.1 & 1530.1 \\
\bottomrule
\end{tabular}
\end{table}

QwenDLM's benchmark-macro clean TPS is $186.88$ for Base and $933.30$ for ORL,
giving approximately $5\times$ with SGLang; with \dlmrt{}, \textsc{ORL} reaches $1{,}298$ tok/s, about $6.9\times$ the SGLang Base, as reported in the main text, and $4.5\times$ the \dlmrt{} Base ($289$ tok/s).
All-output/completed-output TPS ratios are $4.59/4.70$.
At LLaDA's main-table operating points, the clean/all-output/completed-output
ratios are $1.23/1.25/1.29$.
These values follow the request-accounting definitions in
Appendix~\ref{app:eval_details}. The approximately $3\times$ LLaDA result
corresponds to the pair with the same observed mean accuracy in
Figure~\ref{fig:quality_tpf_frontiers} ($2.95\times$ clean TPS).
Qwen Base has one numerically
correct but length-terminated GSM answer; counting it as a failure gives
$1191/1319=90.30\%$ and mean GSM/HE accuracy $86.61\%$ versus $91.03\%$.

\subsection{Speculative-Decoding Comparison}
Table~\ref{tab:spec_context} reconstructs the GSM8K1319/HumanEval164
comparison in Figure~\ref{fig:teaser}b from per-benchmark summaries.
Accuracy and TPS each average the two benchmark-level metrics.
Within each benchmark, TPS pools tokens and request time over each setting's
own clean requests (Equation~\ref{eq:eval-wall-tps}), not a common-clean
intersection across settings.

For the Qwen comparison, matched quality uses the main text's one-sided
criterion: the fastest measured setting per drafter with mean accuracy
at least \textsc{ORL}'s minus $0.5$ percentage points. At \textsc{ORL}'s
$91.03\%$, the corresponding profiles are EAGLE3 QE02 ($94.23\%$, $404.7$ TPS)
and DFlash QD03 ($90.76\%$, $745.5$ TPS); accuracies need not be equal.
Speculative TPF is $1+N_{\mathrm{accepted}}/N_{\mathrm{draft}}$,
not denoising TPF.
\begin{table}[t]
\centering
\small
\setlength{\tabcolsep}{4pt}
\caption{Speculative-decoding results on GSM8K1319 and HumanEval164
(Figure~\ref{fig:teaser}b). Accuracy and clean TPS each average the two
benchmark-level metrics. Length failures count as incorrect and are
excluded from clean TPS; counts cover both tasks. $\tau$ is the
verification threshold: a drafted token $v$ is accepted when the target
model's probability satisfies $p(v)\ge\tau\max_u p(u)$, so $\tau=1$ is
strict verification.
High-failure profiles remain visible, but clean TPS does not charge their
failed-request time.}
\label{tab:spec_context}
\begin{tabular}{llrrrr}
\toprule
Target / method & Profile & $\tau$ & Mean acc. (\%) & Clean TPS & Length failures \\
\midrule
Qwen / AR & QA & \textemdash & 93.40 & 240.1 & 2 \\
Qwen / EAGLE3 & QE01 & 1 & 93.47 & 352.7 & 3 \\
 & QE02 & 0.1 & 94.23 & 404.7 & 3 \\
 & QE03 & 0.03 & 88.39 & 432.3 & 23 \\
 & QE04 & 0.01 & 81.08 & 462.6 & 69 \\
 & QE05 & 0.003 & 65.58 & 483.4 & 220 \\
 & QE06 & 0.001 & 48.63 & 467.6 & 456 \\
Qwen / DFlash & QD01 & 1 & 93.44 & 573.3 & 3 \\
 & QD02 & 0.1 & 92.45 & 690.9 & 3 \\
 & QD03 & 0.03 & 90.76 & 745.5 & 12 \\
 & QD04 & 0.01 & 87.15 & 788.6 & 32 \\
 & QD05 & 0.003 & 80.37 & 827.1 & 75 \\
\midrule
Gemma / AR & GA & \textemdash & 96.76 & 206.8 & 2 \\
Gemma / DFlash & GD01 & 1 & 96.80 & 549.1 & 1 \\
 & GD02 & 0.03 & 97.47 & 605.9 & 2 \\
 & GD03 & 0.003 & 96.83 & 634.0 & 3 \\
 & GD04 & $10^{-4}$ & 95.77 & 639.8 & 10 \\
 & GD05 & $10^{-5}$ & 94.20 & 662.7 & 21 \\
Gemma / EAGLE3 & GE01 & 1 & 96.83 & 331.1 & 0 \\
 & GE02 & 0.03 & 96.91 & 351.3 & 6 \\
 & GE03 & 0.003 & 94.72 & 362.4 & 13 \\
 & GE04 & $10^{-4}$ & 90.43 & 367.0 & 44 \\
 & GE05 & $10^{-5}$ & 84.45 & 385.4 & 111 \\
\bottomrule
\end{tabular}
\end{table}

\paragraph{Configuration.}
Each server uses one H100 and vLLM, serving a single request at a time; two warmup requests
precede measurement. Qwen3-4B-Thinking-2507 uses the drafters
\path{taobao-mnn/Qwen3-4B-Thinking-2507-Eagle} (EAGLE3) and
\path{bingyang-lei/Qwen3-4B-Thinking-Draft-OPD} (DFlash); Gemma-4-26B-A4B-it
uses \path{RedHatAI/gemma-4-26B-A4B-it-speculator.eagle3} (EAGLE3) and
\path{z-lab/gemma-4-26B-A4B-it-DFlash} (DFlash). The Qwen DFlash drafter
targets a different Qwen release. These settings are recorded in
\path{evaluation/configs/paper/speculative.yaml} in the code repository.

\subsection{Additional Model Comparisons}
\begin{figure}[!htbp]
\centering
\includegraphics[width=0.8\linewidth]{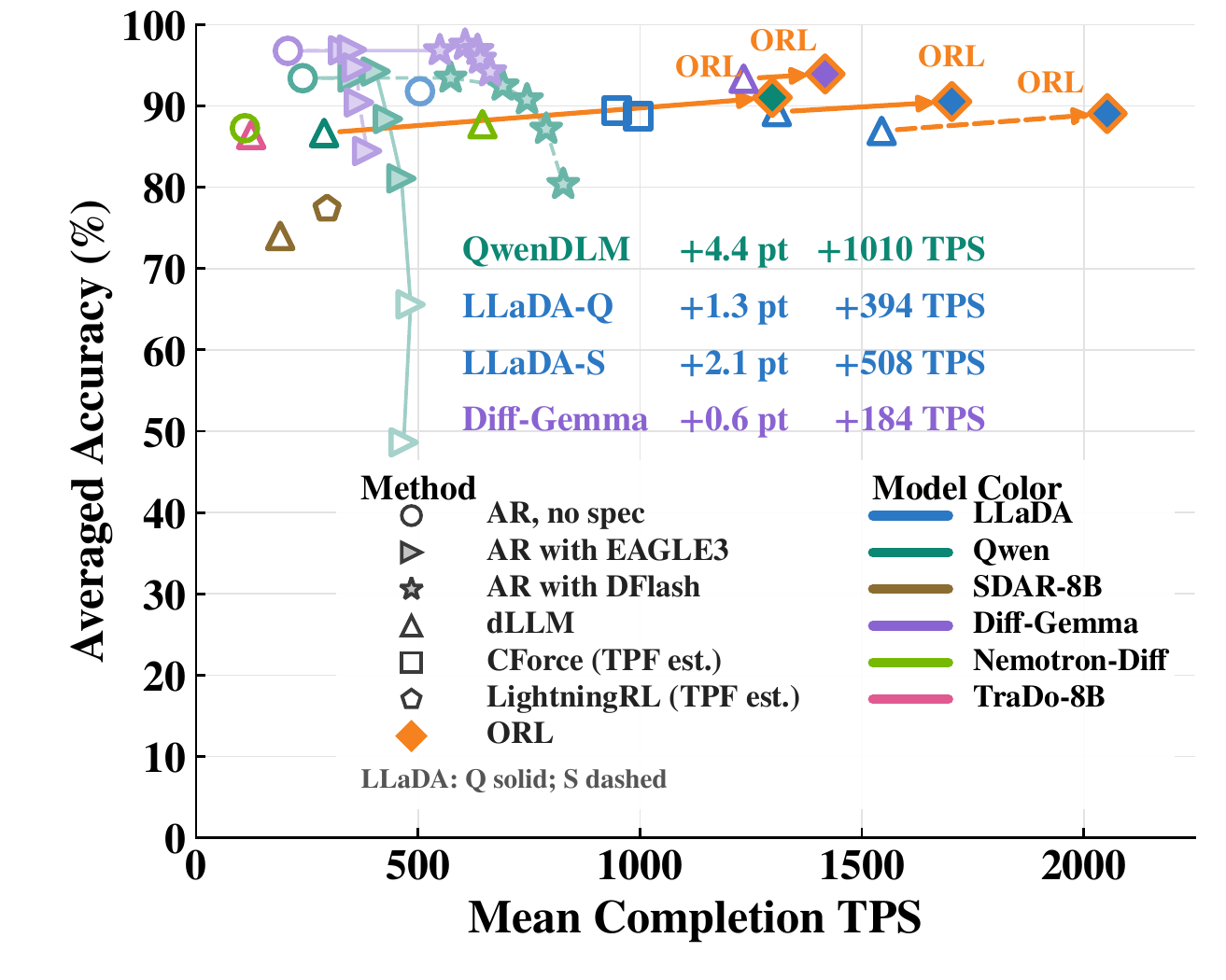}
\caption{\textbf{Accuracy--throughput on GSM8K and HumanEval with additional baselines.}
Both axes average the two benchmark-level metrics. Marker shape denotes
the method and color the model family; orange arrows connect each Base
DLM to its ORL model. LLaDA-Q and LLaDA-S denote Quality and Speed modes.
TPS for CForce, LightningRL, and Nemotron-Diff is estimated as described
above. Base and ORL DLM throughput is measured with \dlmrt{}.}
\label{fig:appendix-gsm-he-baselines}
\end{figure}

Figure~\ref{fig:appendix-gsm-he-baselines} extends the GSM8K/HumanEval
comparison in Figure~\ref{fig:teaser}b with additional model families and
methods, including TraDo-8B~\citep{wang2026revolutionizing}.
CForce~\citep{ren2026cforce} and
Nemotron-Diff~\citep{fu2026nemotron} use published results
converted to TPS; LightningRL~\citep{hu2026lightningrl} uses our measured
TPF converted to TPS (Appendix~\ref{app:baselines}). These converted points
indicate approximate throughput positions. The remaining points use our
measured wall-clock completion throughput for a single request on one H100.

\subsection{Quality--Throughput Sweeps}
\label{app:quality_throughput_frontiers}
\begin{figure}[!t]
\centering
\includegraphics[width=0.82\linewidth]{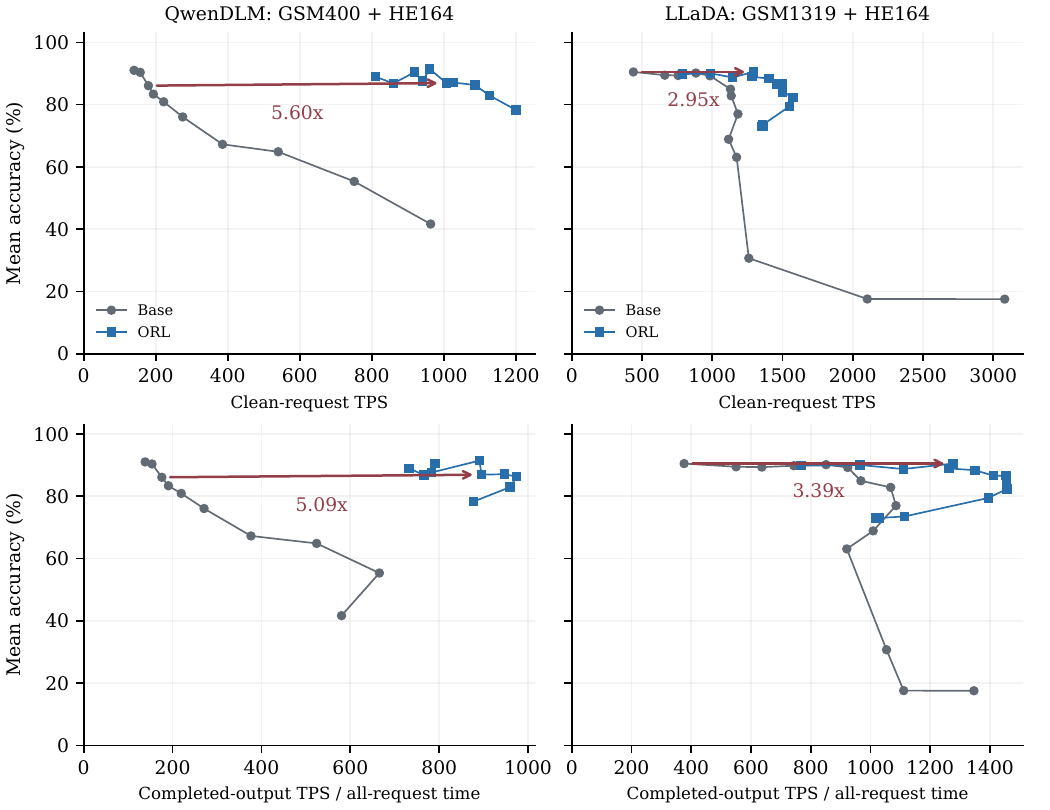}
\caption{Common-task quality--throughput sweeps. Axes average GSM8K and
HumanEval; all settings remain visible. Top: clean TPS; bottom:
completed-output TPS; charging failed-request time. The highlighted Qwen
pair (Base threshold 0.90, ORL entropy budget 0.80) gives $86.09/86.92\%$
mean accuracy and $5.60/5.09\times$ clean/completed TPS gains. The
highlighted LLaDA pair (Base L14, ORL L11) gives $90.49\%$ on both sides,
with $2.95/3.39\times$ gains. Lines connect points in order of increasing
threshold or entropy budget; matching refers to observed mean accuracy.
All settings are measured with SGLang.}
\label{fig:quality_tpf_frontiers}
\end{figure}

Figure~\ref{fig:quality_tpf_frontiers} contains all 48 operating points:
ten each for Qwen Base and ORL, and fourteen per LLaDA checkpoint.
Qwen uses common GSM IDs 0--399 and HE164; LLaDA uses GSM1319 and HE164.
Qwen Base sweeps the acceptance threshold over 0.30, 0.40, 0.50, 0.60,
0.70, 0.80, 0.85, 0.90, 0.95, and 0.99, and Qwen ORL sweeps the entropy
budget over 0.2, 0.3, 0.4, 0.5, 0.6, 0.8, 1.0, 1.5, 2.0, and 3.0; each
LLaDA checkpoint uses profiles L01--L14 (Table~\ref{tab:matched_decoder_grid}).

\subsection{Accuracy--Canvas-TPF Tradeoffs}
\label{app:canvas-tpf-frontiers}
For $K_i$ committed blocks of size $B=32$, define
\begin{equation}
 \operatorname{TPF}^{\mathrm{canvas}}_b
 =\frac{1}{|\mathcal C_b|}\sum_{i\in\mathcal C_b}\frac{BK_i}{D_i}.
 \label{eq:canvas-tpf-noedit}
\end{equation}
Unlike completion TPF, it includes final-block padding; it does not subtract
edits or charge refinement/commit forwards. It is a refinement diagnostic,
not wall TPS. The locked Quality sweep uses the same tasks at all fourteen
settings, separately for HumanEval and GSM8K. At the highlighted HumanEval
pair, both checkpoints pass 146/164 tasks; canvas TPF rises from 4.8395 to
13.9086 ($2.874\times$), while completion TPF rises from 4.3668 to 13.0771
($2.995\times$). Seven tasks pass only each side; accuracy matching is empirical.
Appendix~\ref{app:locked_canvas_validation} gives comparisons at shared settings
and curve validation.

\begin{figure}[!htbp]
\centering
\includegraphics[width=\linewidth]{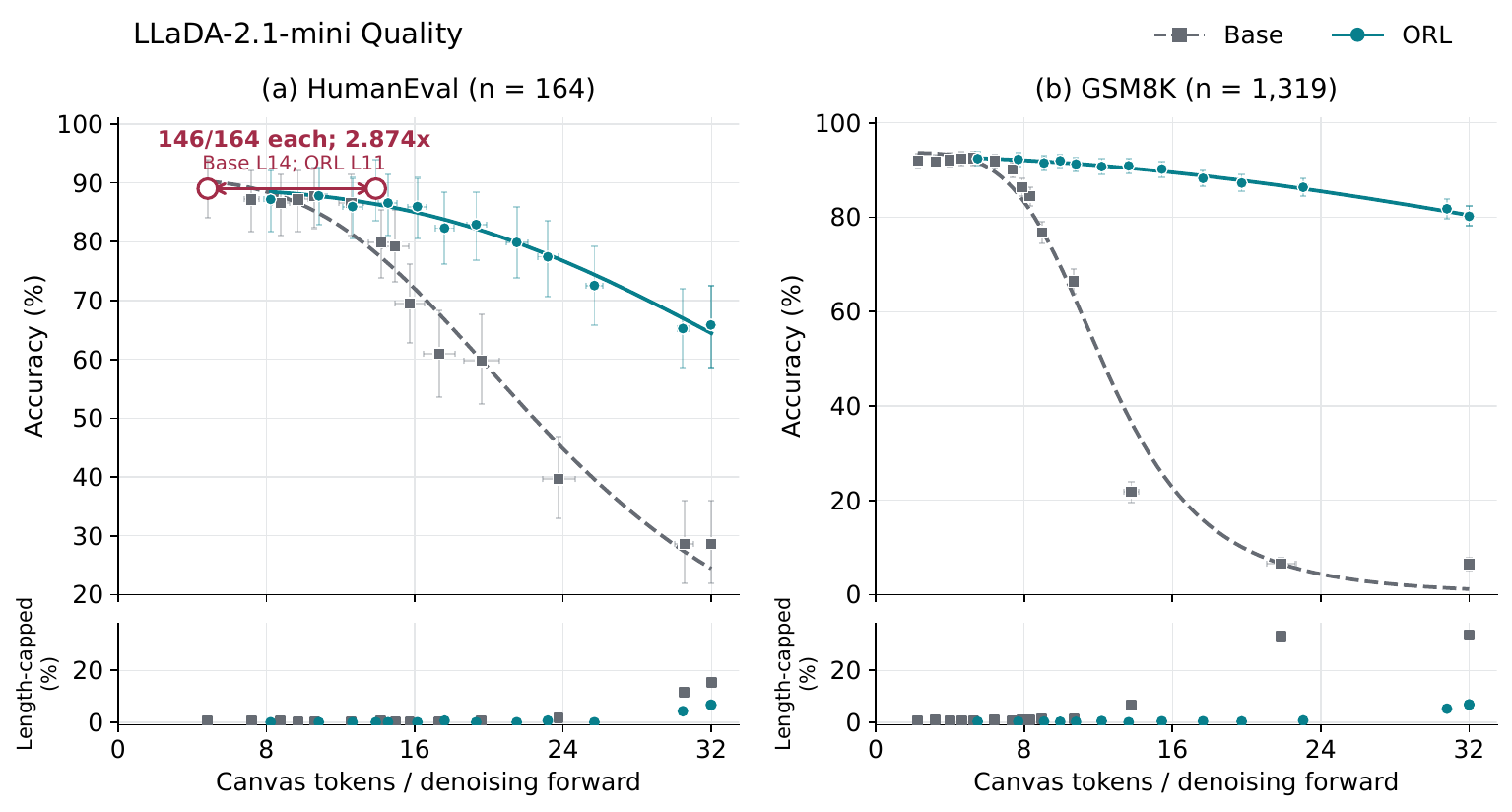}
\caption{LLaDA Quality-mode Base versus ORL, with all operating points.
Lines are independent log-logistic fits within observed ranges; bars are
pointwise 95\% percentile intervals from 2,000 paired task-vector resamples,
conditional on the archived evaluation records. They quantify task-resampling
uncertainty, excluding between-run variation and simultaneous coverage
across operating points. The HumanEval arrow compares
Base profile L14 (threshold 0.99) with ORL profile L11 (threshold 0.85); lower strips
show length failures. Profiles are defined in
Table~\ref{tab:matched_decoder_grid}.}
\label{fig:locked-canvas-tpf-frontiers}
\end{figure}

\section{Additional Ablations}
\subsection{System Cost}
\label{app:system_analysis}
\label{app:phase_breakdown}
\label{app:normalized_efficiency}
At $n=16$ rollouts, packing reduces actor-update time from 368 to 131\,s
and replay materialization from 55 to 16\,s; cost balancing further reduces
the update to 96\,s. Across rollout counts, update cost drops from
0.60--0.93 to 0.24--0.25\,s per 1k supervised tokens. Packing makes each
replay row larger but reduces row count by about $5\times$, so per-row cost
alone is misleading. Asynchronous host staging does not consistently
improve per-unit cost; the large rollout gain requires GPU-resident handoff.

\begin{figure}[!htbp]
\centering
\includegraphics[width=\linewidth]{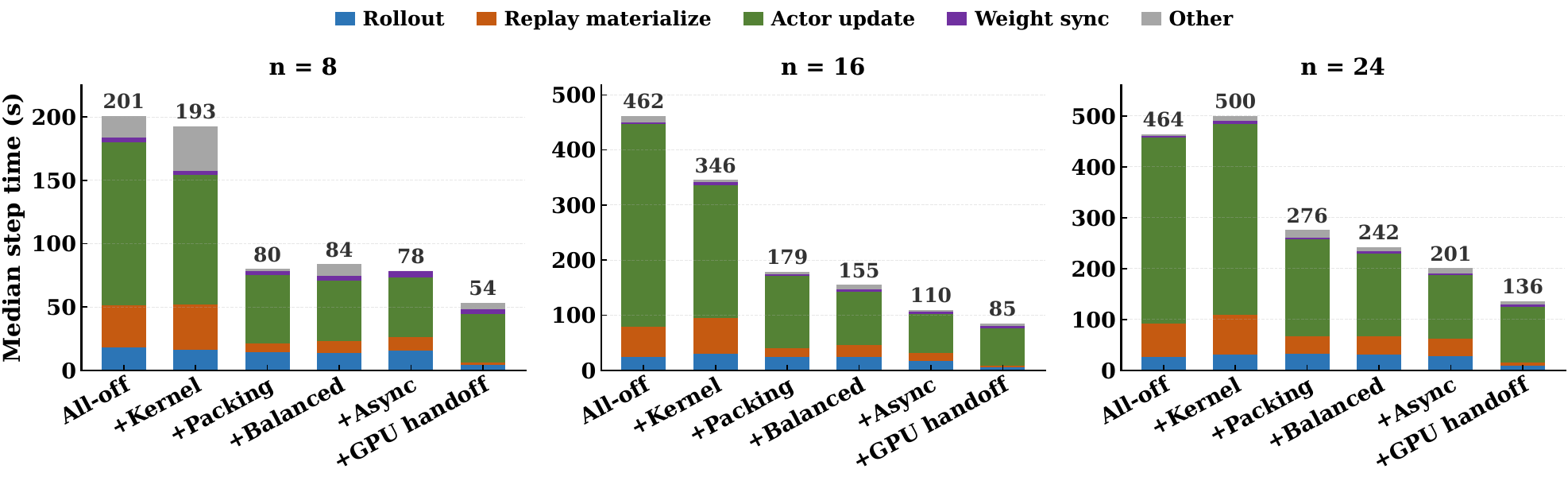}
\includegraphics[width=\linewidth]{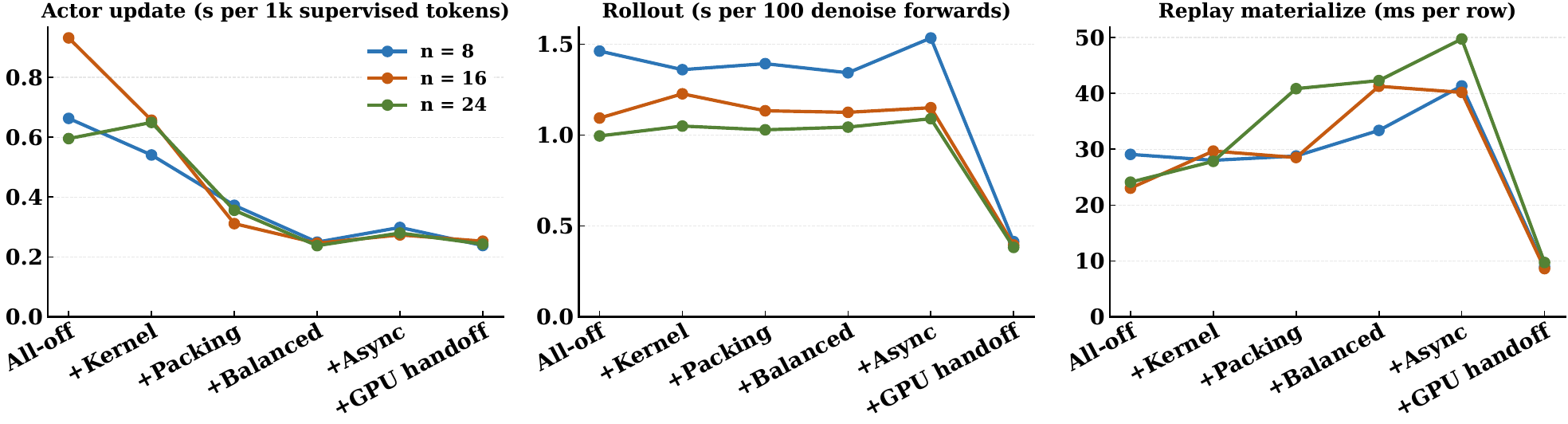}
\caption{QwenDLM system ablation at response cap 2,048. Top: median phase
times; bottom: median costs per 1k supervised tokens, 100 denoising forwards,
and replay row. Medians use post-warmup steps; the main figure instead uses
mean step times. Raw runs can differ in sampled workload.}
\label{fig:app_phase_breakdown}
\label{fig:app_normalized_efficiency}
\end{figure}

\paragraph{Paired handoff and memory.}
\label{app:gpu_handoff_ablation}
\label{app:system_memory}
On identical trajectories at $n=8,16,24$, GPU-resident handoff changes rollout
time from 15.5/17.4/28.4 to 4.2/5.3/9.9\,s and replay materialization from
10.7/14.3/34.1 to 2.0/3.2/5.9\,s, with little actor-update change.
This isolates transport from workload variation.
At cap 2,048, the full system uses about 19\,GB GPU and 113--116\,GB host
memory; baseline host use grows to 381/541/799\,GB. At cap 8,192, the baseline
exceeds the node's 944\,GB host capacity. The plotted 359\,GB system point
disables GPU handoff because the full variant reaches its capture-arena limit;
it does not demonstrate constant host memory at that length.

\begin{figure}[!htbp]
\centering
\begin{subfigure}[t]{0.65\linewidth}
\centering
\includegraphics[width=\linewidth]{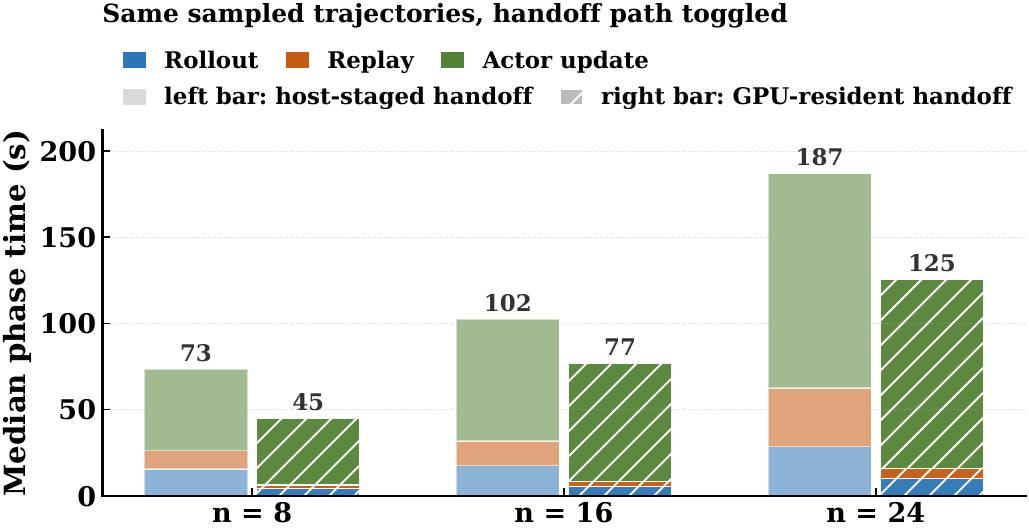}
\caption{Handoff paths on identical trajectories.}
\label{fig:app_async_vs_gpu}
\end{subfigure}
\par\medskip
\begin{subfigure}[t]{\linewidth}
\centering
\includegraphics[width=\linewidth]{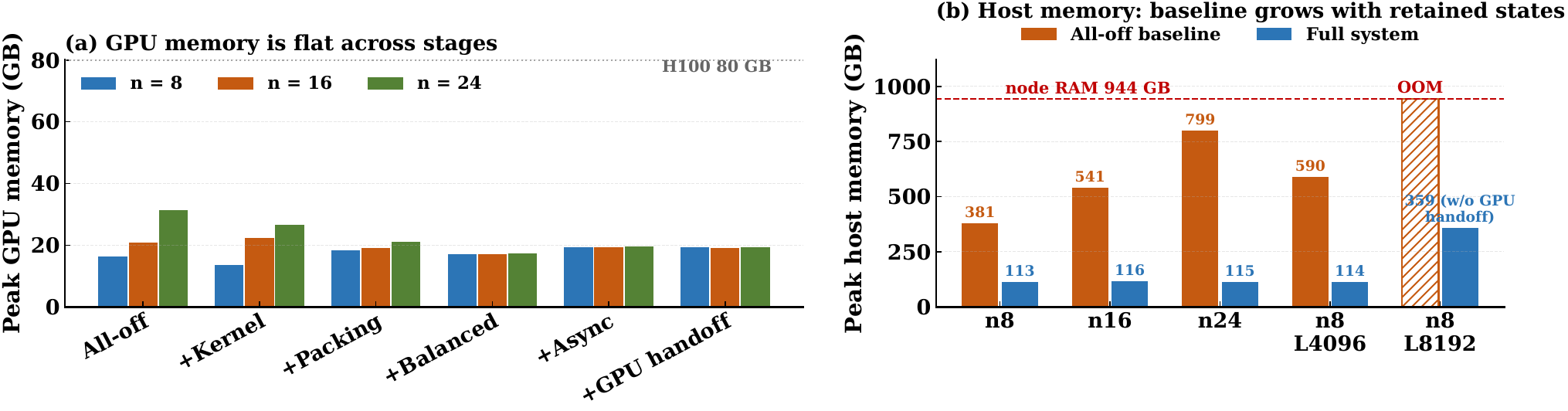}
\caption{GPU and host memory; 8,192-token caveat in text.}
\label{fig:app_memory}
\end{subfigure}
\caption{Transport and memory diagnostics. Paired handoff bars show median
rollout/replay/update times; memory panels show peaks. All GPU configurations
fit an 80\,GB H100.}
\end{figure}

\subsection{Residual Scaling and Block Size}
\label{app:residual_scaling}
Matched QwenDLM/GSM8K runs differ only in $\gamma=8$ residual rescaling.
Within their shared measurement window, excluding skipped updates, the median
$100\|\Delta W\|_2/\|W\|_2$ for the residual MLP is 0.00142\% with scaling
and 0.04431\% without ($31.3\times$); backbone values are
0.00122\%/0.00117\% ($0.96\times$). These are measured optimizer-dependent
update ratios.

\label{app:block_size_mbpp}
The MBPP block-size study complements Figure~\ref{fig:block_scaling}:
ORL raises TPF from 8.6 to 13.0 at block size 32 and from 7.2 to 16.6 at 128.

\begin{figure}[!htbp]
\centering
\begin{subfigure}[t]{0.48\linewidth}
\centering
\includegraphics[width=\linewidth]{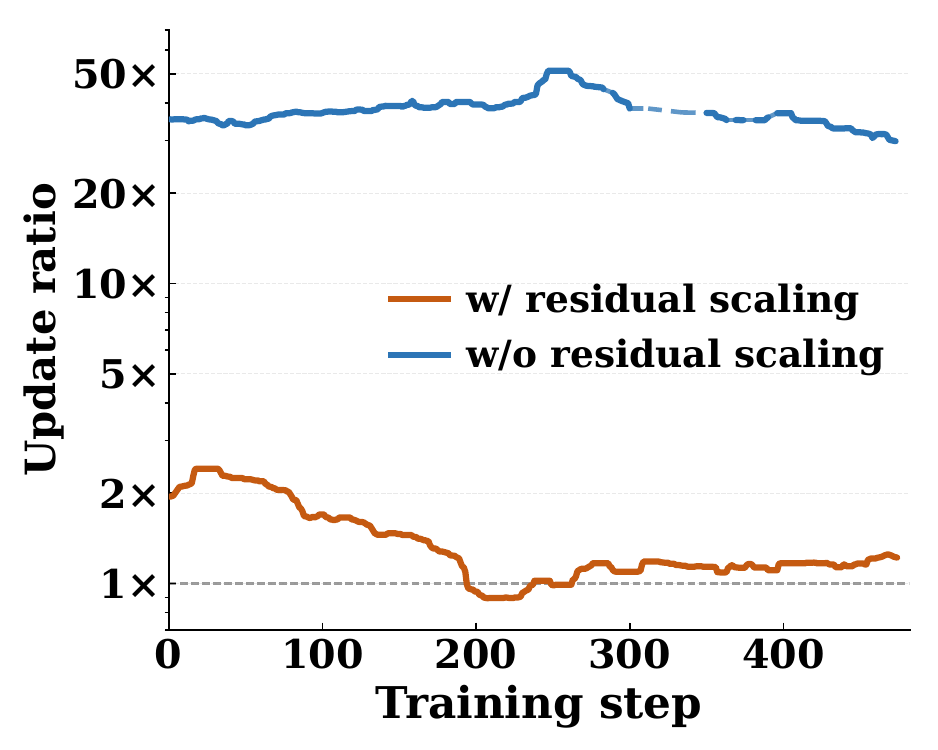}
\caption{Residual scaling controls relative updates.}
\label{fig:alpha_update_ratio}
\end{subfigure}\hfill
\begin{subfigure}[t]{0.48\linewidth}
\centering
\includegraphics[width=\linewidth]{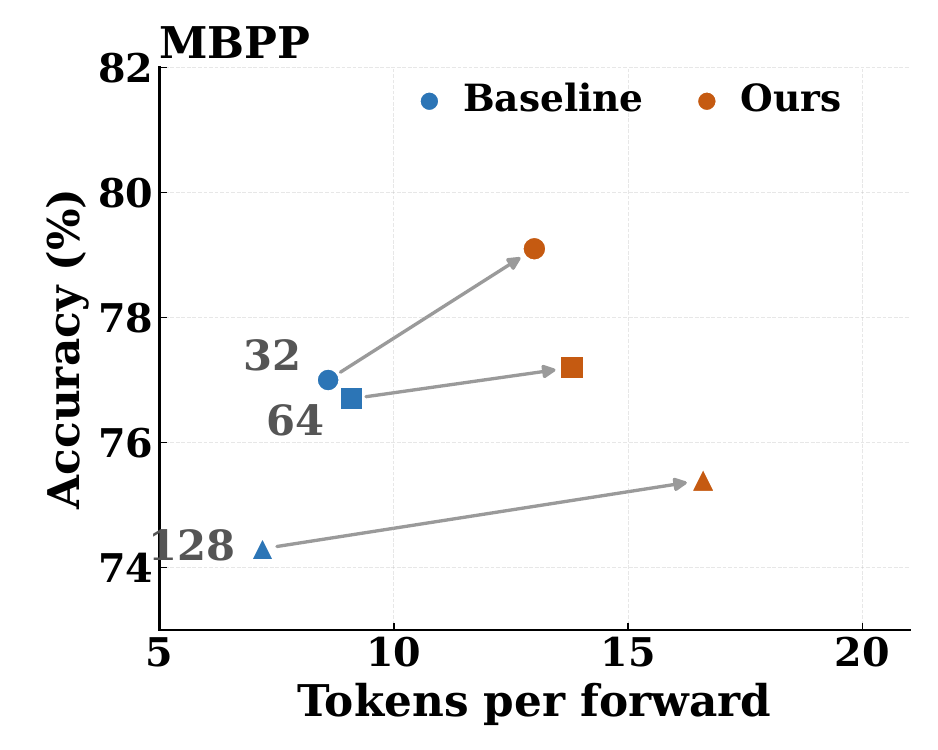}
\caption{LLaDA training block size versus MBPP efficiency.}
\label{fig:app_block_size_mbpp}
\end{subfigure}
\caption{Residual-path stability and block-size sensitivity. Rescaling
primarily changes residual, not backbone, update magnitudes.}
\end{figure}

\begin{figure}[!htbp]
\centering
\includegraphics[width=0.55\linewidth]{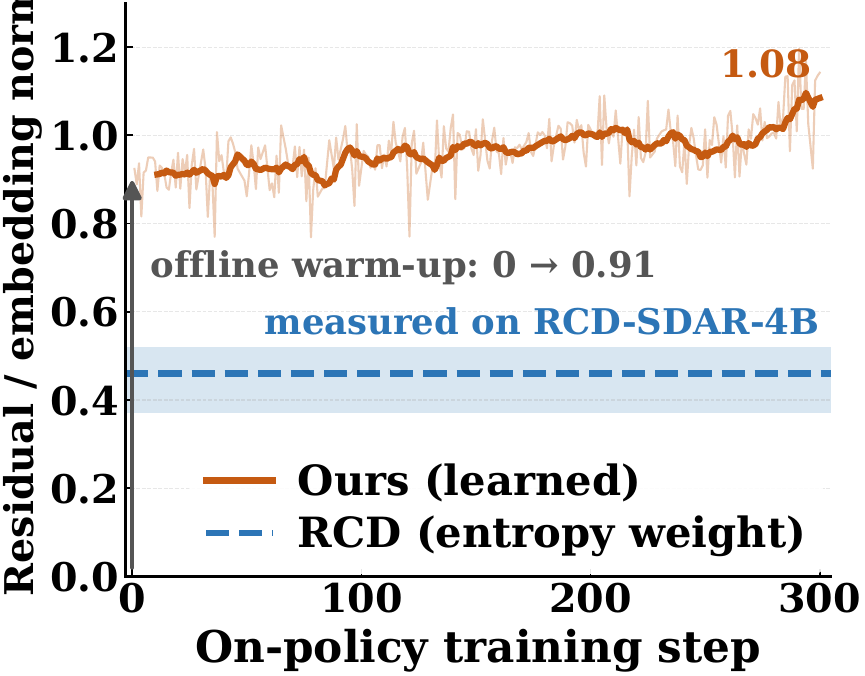}
\caption{QwenDLM residual-to-embedding norm ratio, rising from 0.91 on the
first on-policy rollouts to 1.08 during training. The dashed RCD-SDAR-4B
reference is a separately measured entropy-weighted diagnostic. The
offline-SFT RMS ratio of 1.06 at the same initializer uses a different
population and statistic, so the values are not directly comparable.
These traces describe residual use, not its causal benefit.}
\label{fig:app_residual_norm_ratio}
\end{figure}

\subsection{Training Data and Compute}
\label{app:training-compute}
The data budget counts prompt and response tokens, including failed rollouts,
without multiplying by training epochs. LLaDA training generates
5,000 responses: 0.74M prompt and 3.49M response tokens (4.2M total);
2.37M response tokens pass verification and enter the loss.
Published budgets are about 40M/65M for d3LLM-LLaDA/Dream,
1B for Fast-dLLM-v2, and 54B for SDAR~\citep{qian2026d3llm,wu2026fast,cheng2025sdar}.
dParallel's 40M point is estimated from the same 92,428-trajectory data reused
by d3LLM; counting only 256-token responses gives 23.7M~\citep{chen2026dparallel}.
We compare ORL's budget with post-training recipes, not AR-to-DLM conversion.

Table~\ref{tab:training-compute} counts every allocated H100 along the training
chain leading to each reported checkpoint, including startup, model loading,
rollout, verification, updates, and checkpoint writes. It excludes initializer
production, checkpoint evaluation, other experimental runs, idle gaps, and
attempts failing before their first checkpoint. Including the latter adds about 4.5 GPU-hours for
QwenDLM and at most 8.5 for DiffusionGemma. When a job runs beyond the reported
checkpoint, the next rollout timestamp bounds its endpoint to within one minute.

\begin{table}[!htbp]
\centering
\small
\caption{Post-training compute for the reported checkpoints. Per-job hours
follow chain order; GPU-hours count all allocated H100 80GB GPUs and are rounded.}
\label{tab:training-compute}
\begin{tabular}{lcccc}
\toprule
Model & GPUs & Per job (h) & Total (h) & GPU-hours \\
\midrule
QwenDLM & 8 & $5.00+0.66+1.79$ & 7.45 & 60 \\
LLaDA-2.1-mini & 8 & $3.25+1.27$ & 4.52 & 36 \\
DiffusionGemma & 16 & $8.22+9.55$ & 17.77 & 284 \\
\bottomrule
\end{tabular}
\end{table}

\section{Policy Dependence and Entropy}
\label{app:theory}

\subsection{The Update and Stale Data}
\label{app:theory_onpolicy}
Let $w=(\theta,\phi)$ and let $\nu_w$ be the distribution of captured training
records after verification, selection, and objective-specific weighting.
At iteration $u$, ORL takes the frozen-data gradient
\begin{equation}
 g_u=\left.\nabla_w\mathbb E_{\xi\sim\nu_{w_u}}\ell(w;\xi)\right|_{w=w_u}.
 \label{eq:theory_surrogate}
\end{equation}
Inputs, targets, and source entropies are held fixed for this differentiation.
DiffusionGemma recomputes its source entropy during replay and then detaches
it. Neither recipe differentiates through rollout generation or verification.
In a differentiable common-density model, the full derivative of
$\mathbb E_{\nu_w}\ell(w;\xi)$ would also contain
$\mathbb E_{\nu_w}[\ell(w;\xi)\nabla_w\log\nu_w(\xi)]$.
Thus verified CE is not itself a policy-gradient estimator of task reward.

\paragraph{Stale-gradient bound.}
At a fixed evaluation parameter $w$, assume
$\|\nabla_w\ell(w;\xi)\|_2\le G$ under both record distributions. Then
\begin{equation}
 \left\|\mathbb E_{\nu_w}\nabla\ell(w;\xi)
       -\mathbb E_{\nu_v}\nabla\ell(w;\xi)\right\|_2
 \le 2G\,\operatorname{TV}(\nu_w,\nu_v),
 \qquad \operatorname{TV}(P,Q)=\sup_A|P(A)-Q(A)|.
 \label{eq:theory_stale_bound}
\end{equation}
To prove this, project onto any unit vector and integrate the bounded scalar
gradient against $\nu_w-\nu_v$, whose total variation norm is
$2\operatorname{TV}(\nu_w,\nu_v)$. Taking the supremum proves the inequality.
The gradient bound is an assumption, not automatic for CE.
The result bounds possible mismatch, not harm: different data distributions
can give the same gradient. If $\nu_w\ll\nu_v$, importance weighting can
correct the expectation in principle, subject to support and variance.
Deterministic continuous residuals may instead have disjoint supports.
Fresh collection is therefore a design choice, not a proof that offline
learning is impossible.

\paragraph{Verification and skipped updates.}
With one prompt, $G$ independent rollouts, pass probability $p_w(c)$, and
equal verified-response weights, the expected CE/hybrid gradient per
attempted batch is
$[1-(1-p_w(c))^G]g_w^+(c)$, where $g_w^+(c)$ is the conditional verified
gradient. Indeed, a nonempty verified batch has expected mean $g_w^+(c)$,
and its probability is $1-(1-p_w(c))^G$.
Skipping all-wrong batches therefore reweights prompts. This calculation
does not directly apply to gated AEC's weighted ratio, dependent rollouts,
or nonlinear optimizer updates.

\subsection{Entropy and Token Acceptance}
\label{app:theory_aec}
Use unnormalized entropy $h(p)=-\sum_vp_v\log p_v$; the method uses
$H=h/\log|\mathcal V|$. Suppose proposals are greedy and acceptance uses the
same distribution at fixed threshold $0<\tau<1$. For $n$ eligible positions,
\begin{equation}
 \frac{\#\{i:\max_vp_{iv}\le\tau\}}{n}
 \le\min\left\{1,\frac{\bar h}{-\log\tau}\right\}.
 \label{eq:theory_entropy_acceptance}
\end{equation}
Each failing position contributes at least $-\log\tau$ because
$h(p_i)\ge-\log\max_vp_{iv}$; summing proves the bound.
This concerns confidence, not correctness, and does not directly cover
different confidence temperatures, sampled proposals, entropy budgets,
or re-noising decoders.

If actual rollout means satisfy $H_{k+1}\le rH_k+e_k$ with $0<r<1$,
induction gives $H_k\le r^kH_0+\sum_{j<k}r^{k-1-j}e_j$.
If all unresolved positions are eligible, all above-threshold greedy
proposals commit, and positions never reopen, the remaining count obeys
\begin{equation}
 n_K\le n_0\prod_{k<K}
 \min\left\{1,\frac{H_k\log|\mathcal V|}{-\log\tau}\right\}.
 \label{eq:theory_remaining_positions}
\end{equation}
The vocabulary factor converts normalized entropy to $h$. A right-hand
side below one implies completion, not a correct answer. A finite AEC
penalty on selected training pairs does not impose the assumed contraction
on new rollouts.

Changing masks also changes the measurement. If $U'=U\setminus A$ with
both $A,U'$ nonempty, then
\begin{equation}
 \bar h_{k+1}(U')-\bar h_k(U)
 =[\bar h_{k+1}(U')-\bar h_k(U')]
 +\frac{|A|}{|U|}[\bar h_k(U')-\bar h_k(A)].
 \label{eq:theory_cohort_decomposition}
\end{equation}
Removing confident positions can raise the remaining mean without worsening
any prediction. Conversely, temperature sharpening lowers entropy:
$d h(\operatorname{softmax}(\beta z))/d\beta
=-\beta\operatorname{Var}_{p_\beta}(z)\le0$, while leaving a wrong argmax
wrong. These are reasons to measure quality separately.

\subsection{Optimizer and Rescaling}
\label{app:theory_entropy_update}
At a fixed captured input and mask, $p=\operatorname{softmax}(z)$ gives
\begin{equation}
 \Delta h=-\operatorname{Cov}_{v\sim p}(\log p_v,\Delta z_v)
          +O(\|\Delta z\|_2^2).
 \label{eq:theory_entropy_update}
\end{equation}
This follows from $\partial h/\partial z_v=-p_v(\log p_v+h)$.
The actual $\Delta z$ depends on the network Jacobian, all losses,
clipping, and optimizer state; a policy-gradient covariance specialization
need not describe CE/AEC/AdamW. Rollout-average changes additionally include
state-distribution shift.

\label{app:theory_rescaling}
For a compensated projection $\widetilde W=\gamma W$ evaluated as
$\widetilde W/\gamma$, the chain rule gives
$\nabla_{\widetilde W}L=\nabla_WL/\gamma$.
Equal-rate SGD reduces the effective $W$ update by $\gamma^2$.
Adam with consistently scaled moments and negligible epsilon reduces it
by approximately $\gamma$, not $\gamma^2$; clipping, untransformed moments,
and epsilon can change this relation. AdamW's relative decay is unchanged.
Equation~\ref{eq:residual_rescaling} proves forward-function preservation;
stability and the measured update ratios remain empirical.

\section{Accuracy--TPF Curves}
\label{app:tpf_curve_analysis}
\label{app:theory_tpf}

\subsection{A Descriptive Model}
For an operating point $\tau$, write $S=\mathbf1\{\text{not length-terminated}\}$,
$Z$ for benchmark correctness, and $U=C/D$. The sweep estimates
$a(\tau)=\mathbb E[SZ]$ and $t(\tau)=\mathbb E[U\mid S=1]$, not
$P(Z=1\mid U=t)$. In particular,
$a=(1-P(S=0))P(Z=1\mid S=1)$; collapse changes both axes.
For the canvas diagnostic, replace $C$ by $BK$ throughout.

We use the log-logistic family
\begin{equation}
 a(t)=\frac{A}{1+(t/f_0)^\kappa},\qquad
 0<A\le1,\quad f_0,\kappa>0.
 \label{eq:curve_family}
\end{equation}
One sufficient latent model assumes a solvable fraction $A$, sensitivity
$\Lambda\sim\operatorname{Exp}(\eta)$ within that fraction, and retention
$\exp(-\Lambda t^\kappa)$. Then
\[
 A\int_0^\infty e^{-\lambda t^\kappa}\eta e^{-\eta\lambda}\,d\lambda
 =\frac{A\eta}{\eta+t^\kappa},
 \qquad f_0=\eta^{1/\kappa}.
\]
Exponential sensitivity and power exposure are assumptions, not consequences
of residual decoding. A gamma sensitivity instead gives
$A(1+t^\kappa/\eta)^{-r}$. Neither identity explains why an aggregate,
termination-conditioned TPF should equal the assumed exposure.

The curve has plateau $A$, half-plateau location $f_0$, and
$d a/d\log t|_{t=f_0}=-A\kappa/4$. At relative retention $q$,
\begin{equation}
 t_q=f_0\left(\frac{1-q}{q}\right)^{1/\kappa},\qquad
 \frac{t_i(q)}{t_j(q)}
 =\frac{f_i}{f_j}
 \left(\frac{1-q}{q}\right)^{1/\kappa_i-1/\kappa_j}.
 \label{eq:theory_quality_factor}
\end{equation}
Thus a quality-independent horizontal factor requires shared $\kappa$.
At common absolute accuracy $\alpha$,
$t_m(\alpha)=f_m(A_m/\alpha-1)^{1/\kappa_m}$.
The ratio is constant on an open accuracy interval only if both plateaus
and shapes match: differentiating its logarithm gives
\[
 \frac{d\log(t_R/t_B)}{d\alpha}
 =\frac{A_B}{\kappa_B\alpha(A_B-\alpha)}
  -\frac{A_R}{\kappa_R\alpha(A_R-\alpha)};
\]
equating polynomial coefficients after cross-multiplication requires
$\kappa_B=\kappa_R$ and $A_B=A_R$.
Otherwise speed ratios must name their accuracy level.
When each block requires a denoising forward, canvas TPF is at most $B$;
$f_0>B$ is an extrapolated parameter, not an attainable operating point.
Even $t_{0.9}$ requires a measured-range check.

\subsection{Locked LLaDA Sweep}
\label{app:locked_canvas_validation}
All 56 Base/ORL points are reconstructed from 41,524 request records:
fourteen Quality-mode profiles on HE164 and GSM1319 for each checkpoint.
Task IDs match across profiles/checkpoints; accuracy, clean counts,
and both TPF definitions reproduce saved aggregates within $10^{-10}$.
HumanEval uses base tests and GSM8K its original extractor.

\paragraph{Observed operating points.}
The HumanEval pair in Figure~\ref{fig:locked-canvas-tpf-frontiers} has a
$2.874\times$ canvas-TPF ratio at 146/164 correct on both sides.
Recomputing on the 163 tasks clean for both gives $2.883\times$.
For 20,000 paired task bootstraps, the canvas-ratio interval is
$[2.737,3.012]$ and the accuracy-difference interval is
$[-4.268,4.268]$ percentage points. These pointwise intervals condition
on the archived evaluation records and the displayed pair; they quantify
task-resampling uncertainty, without estimating between-run variability or
accounting for comparisons across the full sweep. They do not establish
population equivalence.
On GSM8K, maximizing observed canvas TPF subject
to empirical accuracy floors of 90\%/92\% gives ratios 2.093/1.461.
The resulting operating points need not have identical accuracies.

\paragraph{Shared evaluation profiles.}
Table~\ref{tab:matched_decoder_grid} compares Base and ORL under identical
settings and retains every operating point, including quality losses.
All profiles use block size 32, edit threshold 0.5, sixteen refinement
steps, and confidence temperature 1.0, and differ only in the acceptance
threshold. This comparison does not isolate
AEC from the other training components.

\begin{table}[!htbp]
\centering
\small
\caption{LLaDA Base--ORL comparisons across all fourteen shared profiles.
Threshold: minimum top-token probability for accepting a position.
$\Delta a$: ORL minus Base accuracy in percentage points.
$R$: clean \emph{completion}-TPF ratio, not padded canvas or wall TPS.}
\label{tab:matched_decoder_grid}
\begin{tabular}{crrrrr}
\toprule
 & & \multicolumn{2}{c}{HumanEval} & \multicolumn{2}{c}{GSM8K} \\
\cmidrule(lr){3-4}\cmidrule(lr){5-6}
Profile & Threshold & $\Delta a$ & $R$ & $\Delta a$ & $R$ \\
\midrule
L01 & 0.00 & +37.20 & 1.048 & +73.77 & 0.920 \\
L02 & 0.05 & +37.20 & 1.096 & +73.69 & 1.380 \\
L03 & 0.10 & +25.61 & 1.344 & +60.05 & 2.190 \\
L04 & 0.20 & +12.80 & 1.368 & +19.94 & 2.111 \\
L05 & 0.30 & +16.46 & 1.392 & +10.46 & 2.159 \\
L06 & 0.40 & +10.37 & 1.423 & +3.79 & 2.077 \\
L07 & 0.50 & +3.66 & 1.352 & +3.79 & 1.914 \\
L08 & 0.60 & +2.44 & 1.296 & +0.83 & 1.812 \\
L09 & 0.70 & -0.61 & 1.346 & -1.14 & 1.862 \\
L10 & 0.80 & -1.22 & 1.432 & -1.21 & 2.011 \\
L11 & 0.85 & +1.83 & 1.500 & -0.45 & 2.110 \\
L12 & 0.90 & -0.61 & 1.508 & -0.61 & 2.235 \\
L13 & 0.95 & +0.61 & 1.574 & +0.53 & 2.334 \\
L14 & 0.99 & -1.83 & 1.774 & +0.45 & 2.379 \\
\bottomrule
\end{tabular}
\end{table}

At profile L11, HE accuracy is 143/164 versus 146/164 and completion TPF is
8.720 versus 13.077; GSM accuracy is 1219/1319 versus 1213/1319, with TPF
4.045 versus 8.535. Across 20,000 joint task-vector bootstraps (seed 20260925),
a max-standardized-deviation band over 56 accuracy/log-ratio coordinates
gives TPF-ratio intervals [1.385,1.623]/[2.059,2.163] and accuracy-difference
intervals [-6.00,9.66]/[-2.74,1.83] for HE/GSM. These approximate familywise
intervals cover the displayed profiles and benchmarks conditional on the
evaluated checkpoints and archived request records. They quantify
task-resampling uncertainty without estimating variation across independent
training runs or repeated evaluations.

\begin{table}[!htbp]
\centering
\small
\caption{Conditional accuracy-prediction RMSE (percentage points), including
endpoint folds. This holds out operating points, not tasks or seeds.}
\label{tab:locked_canvas_cv}
\begin{tabular}{lrrrr}
\toprule
 & \multicolumn{2}{c}{One point out}
 & \multicolumn{2}{c}{Adjacent pair out} \\
\cmidrule(lr){2-3}\cmidrule(lr){4-5}
 & Log-logistic & Isotonic & Log-logistic & Isotonic \\
\midrule
HE / Base & 4.46 & 3.66 & 6.65 & 5.76 \\
HE / ORL & 1.69 & 1.42 & 2.39 & 1.31 \\
GSM / Base & 6.91 & 8.28 & 7.44 & 9.66 \\
GSM / ORL & 0.46 & 0.44 & 0.59 & 0.89 \\
\bottomrule
\end{tabular}
\end{table}

\paragraph{Held-out operating points.}
We fit each checkpoint/benchmark independently with a marginal binomial
working loss, four initializations, and bounds $A\in[.001,1]$,
$f_0\in[.1,500]$, $\kappa\in[.1,50]$. Each fold excludes one operating point or
one adjacent pair in the ordered grid, predicting all fourteen points once.
Both predictors receive held-out measured canvas TPF. The comparator is
decreasing isotonic regression with linear interpolation, endpoint clamping,
and pooled weights at duplicate coordinates.

The log-logistic fit is not uniformly better. In 400 paired task-vector
bootstrap refits, 95\% intervals for isotonic minus log-logistic
adjacent-pair RMSE are [-4.18,3.00]/[-2.04,0.22] for HE Base/ORL and
[1.10,3.30]/[-0.06,0.37] for GSM Base/ORL. Only GSM Base's individual
interval excludes zero in favor of the curve; these intervals are
exploratory and not simultaneous.
Full-data Base/ORL shape estimates are 3.39/3.02 on HE and 4.64/2.02 on
GSM, not evidence of a shared shape. ORL's half-plateau estimates 44.03/80.39
exceed the canvas bound 32. The curve is a compact empirical approximation,
not an identified latent-difficulty law or proof of an AEC mechanism.

\end{document}